\pdfoutput=1
\documentclass[11pt]{article}

\usepackage[final]{acl}

\usepackage{times}
\usepackage{latexsym}
\usepackage{url}
\usepackage{tcolorbox}

\usepackage{booktabs}       % professional-quality tables
\usepackage{nicefrac}       % compact symbols for 1/2, etc.
\usepackage{microtype}      % microtypography
\usepackage{subfigure}
\usepackage{csquotes}
\usepackage{amsthm,amssymb,amsfonts,amsmath}
\usepackage{caption,subfigure,graphicx,epstopdf,multirow,multicol,booktabs,verbatim,wrapfig,bm}
\usepackage{xspace}
\usepackage{makecell}
\theoremstyle{definition}
\usepackage{pifont}

\usepackage[colorinlistoftodos,prependcaption]{todonotes}
\usepackage{xargs}
\usepackage{xcolor}
\usepackage{threeparttable}
\definecolor{BrickRed}{RGB}{203,65,84}
\definecolor{OliveGreen}{RGB}{85,107,47}

\newcommandx{\cmt}[2][1=Comment]{\vspace{2pt}\todo[inline,backgroundcolor=black!5]{\textbf{(#1)} \; #2}}

\newcommand{\model}{\text{iAgent}}
\newcommand{\modelplus}{\text{i$^2$Agent}}
\newcommand{\dataset}{\textsc{InstructRec}\xspace}

\newcommand{\cmark}{\color{OliveGreen}\ding{51}}%
\newcommand{\xmark}{\color{BrickRed}\ding{55}}%

\definecolor{softblue}{RGB}{100, 149, 237}

\definecolor{PromptTemplateColor}{RGB}{0, 102, 204} % 深蓝色
\definecolor{KeyWordsColor}{RGB}{0, 153, 51} % 深绿色
\definecolor{HighlightReplyColor}{RGB}{204, 102, 0} % 深橙色
\newcommand{\template}[1]{\textcolor{PromptTemplateColor}{#1}}
\newcommand{\keywords}[1]{\textcolor{KeyWordsColor}{#1}}
\newcommand{\highlightreply}[1]{\textcolor{HighlightReplyColor}{#1}}

\usepackage[T1]{fontenc}
\usepackage[utf8]{inputenc}

\usepackage{microtype}

\usepackage{inconsolata}

\usepackage{graphicx}

\title{$\model$: LLM Agent as a Shield \\ between User and Recommender Systems}

\author{
 \textbf{Wujiang Xu\textsuperscript{1}},
 \textbf{Yunxiao Shi\textsuperscript{2}},
 \textbf{Zujie Liang\textsuperscript{3}},
 \textbf{Xuying Ning\textsuperscript{4}},
 \textbf{Kai Mei\textsuperscript{1}},
\\
 \textbf{Kun Wang\textsuperscript{5}},
 \textbf{Xi Zhu\textsuperscript{1}},
 \textbf{Min Xu\textsuperscript{2}},
 \textbf{Yongfeng Zhang\textsuperscript{1}}
\\
 \textsuperscript{1}Rutgers University,
 \textsuperscript{2}University of Technology Sydney,
 \textsuperscript{3}Independent Researcher,\\
 \textsuperscript{4}University of Illinois Urbana-Champaign,
 \textsuperscript{5}Nanyang Technological University
\\
\small\texttt{{\href{mailto:wujiang.xu@rutgers.edu}{wujiang.xu@rutgers.edu}, \href{mailto:yongfeng.zhang@rutgers.edu}{yongfeng.zhang@rutgers.edu}
 }}
}

\begin{document}
\maketitle
\begin{abstract}
Traditional recommender systems usually take the user-platform paradigm, where users are directly exposed under the control of the platform's recommendation algorithms.
However, the defect of recommendation algorithms may put users in very vulnerable positions under this paradigm. 
First, many sophisticated models are often designed with commercial objectives in mind, focusing on the platform's benefits, which may hinder their ability to protect and capture users' true interests. Second, these models are typically optimized using data from all users, which may overlook individual user's preferences. 
Due to these shortcomings, users may experience several disadvantages under the traditional user-platform direct exposure paradigm, such as lack of control over the recommender system, potential manipulation by the platform, echo chamber effects, or lack of personalization for less active users due to the dominance of active users during collaborative learning.
Therefore, there is an urgent need to develop a new paradigm to protect user interests and alleviate these issues.
Recently, some researchers have introduced LLM agents to simulate user behaviors, these approaches primarily aim to optimize platform-side performance, leaving core issues in recommender systems unresolved. 
To address these limitations, we propose a new user-agent-platform paradigm, where agent serves as the protective shield between user and recommender system that enables indirect exposure. To this end, we first construct four recommendation datasets, denoted as $\dataset$, along with user instructions for each record. 
To understand user's intention, we design an Instruction-aware Agent ($\model$) capable of using tools to acquire knowledge from external environments. Moreover, we introduce an Individual Instruction-aware Agent ($\modelplus$), which incorporates a dynamic memory mechanism to optimize from individual feedback. Results on four $\dataset$ datasets demonstrate that $\modelplus$ consistently achieves an average improvement of 16.6\% over SOTA baselines across ranking metrics. Moreover, $\modelplus$ mitigates echo chamber effects and effectively alleviates the model bias in disadvantaged users (less-active), serving as a shield between user and recommender systems. 
Datasets and code are publicly available at the URL\footnote{\url{https://github.com/agiresearch/iAgent}}.
\end{abstract}

\section{Introduction}
Over the past decades, recommender systems have been extensively applied across various platforms to provide personalized services to users. In the traditional ecosystem of recommender systems, the recommendation models are predominantly delivered through a user-platform paradigm, where users are directly subject to the platform's algorithms.
% , often without adequate algorithms protection. 
This paradigm places users in a vulnerable position, such as lack of control over their recommendation results, potentially being
manipulated by the platform's recommendation algorithms, being trapped in echo chambers, or lack of personalization
for those less active users due to the active users' dominance of the recommendation algorithm.
% as their personalized preferences are frequently overlooked, constrained by the limitations inherent to this paradigm.

\begin{figure}[tb!]
\centering
% 第一行的两个子图
\subfigure[]{
\begin{minipage}[t]{0.60\linewidth}
\centering
\includegraphics[width=\linewidth]{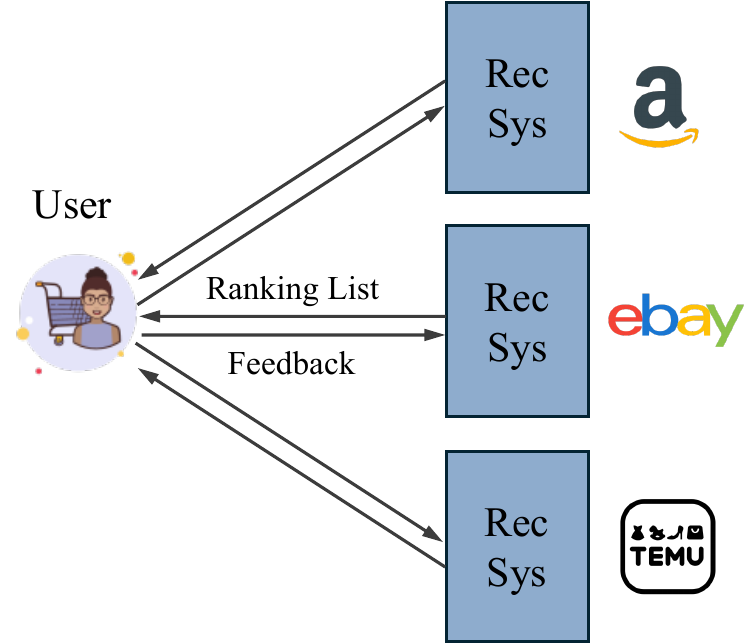}
%\caption{fig1}
\end{minipage}%
}% 
\\
\subfigure[]{
\begin{minipage}[t]{0.80\linewidth}
\centering
\includegraphics[width=\linewidth]{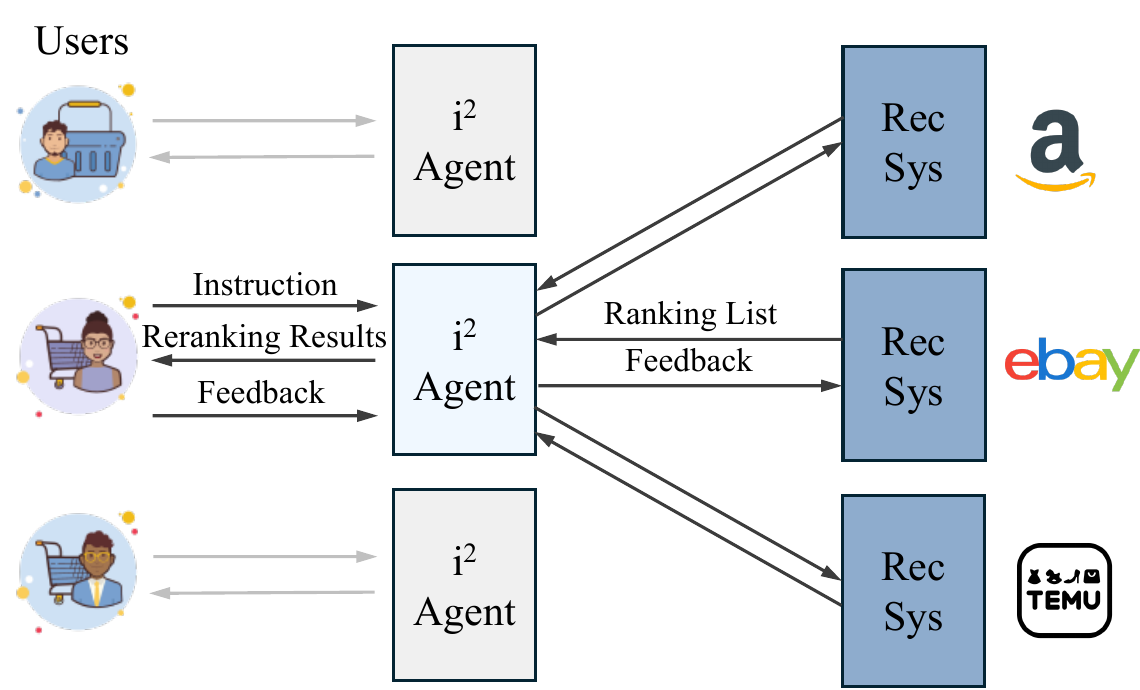}
%\caption{fig2}
\end{minipage}%
}% 
\\ % 开始新的一行
\centering
\captionsetup{font={small}}
\vspace{-1.0em}
\caption{(a) Previous recommendation ecosystem primarily focused on designing sophisticated models to enhance the ranking performance so as to increase platform's benefit. However, they overlooked the user's proactive instructions and put users under the direct control of recommender systems. (b) In contrast, we build an individual instruction-aware agent for each user, which generates re-ranking results based on the user's active instructions. The agent's memory component is influenced solely by the individual user, providing an individualized personal service.
}
\label{fig:intro}
\vspace{-18pt}
\end{figure}

Firstly, the majority of recommendation models~\citep{wideanddeep,sasrec,gru4rec} are designed to optimize the commercial objectives of the platforms, such as increasing user clicks or conversion rates in e-commerce. This often results in users losing sight of their actual needs due to the algorithmic manipulation~\citep{nopersonalized,edizel2020fairecsys,manupilate}.
Secondly, although recommendation models aims at offering personalized services, they are primarily optimized based on data from all users, paying insufficient attention to individual preferences and unique interests~\citep{patro2020fairrec,fairnessrec,ge2022explainable}.
As a consequence of these shortcomings, users often fall into the echo chamber effects~\citep{echochamber,echoeffect2,echoeffect1}, where algorithms reinforce user's existing interests or beliefs through repeated recommendation of homogeneous items, leading to a lack of diversity in recommended contents. Furthermore, the models tend to be biased towards advantaged (active) users, neglecting the interests of disadvantaged (less-active) users, resulting in a lack of personalization for some users.

% \textbf{multi-interest network have been proposed to tackle these problems, but lose the confidence of .}
% first problem are not resolved. first describe the second problem. and describe details of these methods.
To tackle these issues, researchers have approached the problem from various perspectives. On one hand, efforts are made to better understand user interests, such as using user's explicit feedback to improve the model performance and explanation~\citep{zhang2014explicit,xie2021deep} or allowing users to better express their needs through conversational recommender sysetms (CRS) \citep{CRSsurvey,zhang2018CRS}. 
% phrase users' instructions to explore the attributes of specific interest items. 
On the other hand, comprehensive models are developed to infer user interests from various dimensions, such as capturing user's diverse interests based on multi-behavior and multi-interest modeling~\citep{DIN,DIEN,MIND}.
Most recently, language-based agents 
% equipped with tool learning 
are utilized to mock the user behaviors and explore the user interests~\citep{agentcf,generativerecagent}.

However, the two challenges remain insufficiently addressed due to the reliance on modeling user interests across all users' data and the focus on platform-side optimization. \emph{To address these limitations, we propose a new user-agent-platform paradigm, where agent serves as the protective shield between user and recommender system that enables indirect exposure.} Our contributions are three-fold:

% In this paper, we construct four datasets with user instruction via LLM-based on the existing recommendation datasets~\citep{amazon18dataset,steamdataset}. To resolve these challenges, we design a instruction-aware agent for each user, which can learn open-world knowledge and rerank the recommendation results from the platforms. 
% % It caters to our key motivation: \textit{We aim at designing personalized agent algorithms for each user, acting as a shield to protect them from being manipulated by platform-side algorithms.}
% Moreover, to enhance the personalized ability of agent, we design dynamic user-specific memory mechanism. It construct the dynamic profile according to user's individual feedback and extract the dynamic interest based on the instruction.
% Different from existing recommendation models, our $\modelplus$ is optimized specifically for individual users and is not influenced by the interests or behaviors of other users.
% Our key contribution can be summarized as follows.

$\circ$\; \emph{New Datasets and Problem}: To provide benchmarks for the new user-agent-platform paradigm, we construct four recommendation datasets with user-driven instructions, referred to as $\dataset$, constructed from existing datasets such as Amazon, Goodreads, and Yelp. Building on this, we propose an Instruction-aware Agent ($\model$), designed to learn user interests from the provided free-text instructions while leveraging external knowledge to act as a domain-specific expert. Unlike the instructions in CRS~\citep{2018CRS} and Webshop~\citep{webshop}, the free-text instructions in $\dataset$ allow users to flexibly express their requirements
% reflecting the user's generalized interests 
beyond just product attributes. We provide problem definition in Appendix~\ref{sec:definitions}.
% , enabling users to better control the recommender systems based on natural language instructions.

$\circ$\; \emph{Agent Learning from Individual Feedback}: We design Individual Instruction-aware Agent ($\modelplus$), incorporating a dynamic memory mechanism with a profile generator and dynamic extractor to further explore user interests and learn from user's individual feedback. The profile generator constructs and maintains a user-specific profile by leveraging historical information and feedback. The dynamic extractor captures evolving profiles and interests based on the user's real-time instructions. Different from existing recommendation models, $\modelplus$ is optimized specifically for individual users and is not influenced by the interests or behaviors of other users, protecting the interests of less-active users. 

$\circ$\; \emph{Empirical Results}: Empirical experiments on four datasets demonstrate that our $\modelplus$ consistently outperforms state-of-the-art approaches, achieving an improvement of up to 16.6\% on average across standard ranking metrics.
Besides, we evaluate the impact of the echo chamber effect as well as the performance of both active and less-active users separately. From the overall empirical results, it validates that our proposed $\modelplus$ serve as a shield between user and recommender systems. 

% \subsection{Problem Definition} 

% \subsection{Dataset Construction}
% move to experiment?

\section{Methodology}
\begin{figure*}[tb!]
\centering
% 第一行的两个子图
\subfigure[$\model$]{
\begin{minipage}[t]{0.95\linewidth}
\centering
\includegraphics[width=\linewidth]{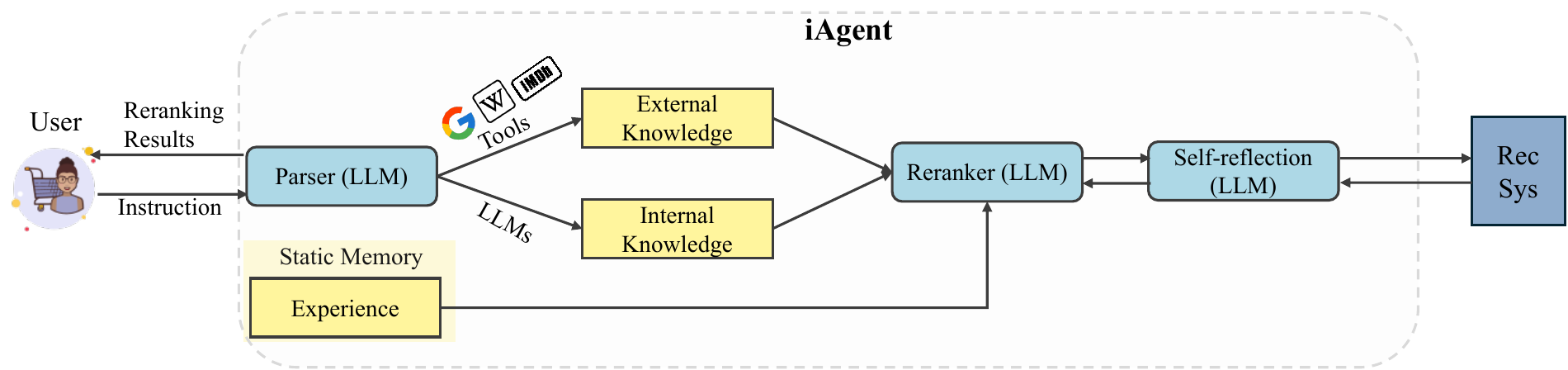}
%\caption{fig1}
\end{minipage}%
}%
\hfill % 填充所有可用的水平空间
\\
\subfigure[$\modelplus$]{
\begin{minipage}[t]{0.95\linewidth}
\centering
\includegraphics[width=\linewidth]{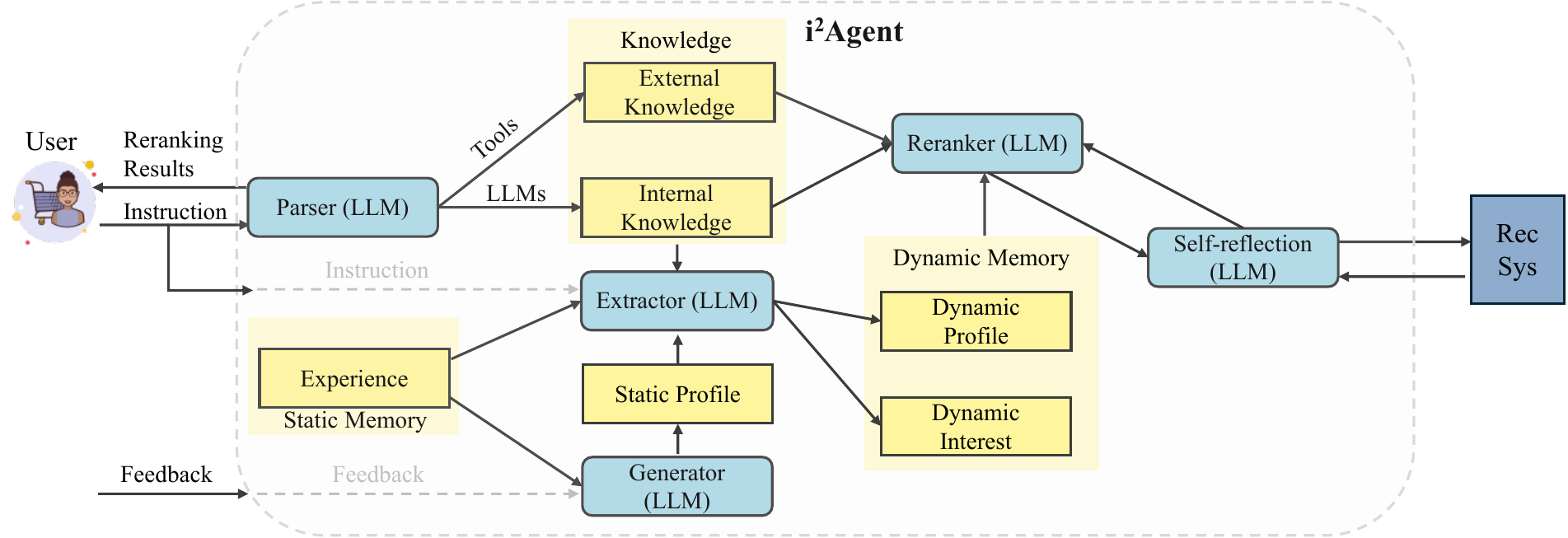}
%\caption{fig2}
\end{minipage}%
}% 
\\ % 开始新的一行
\centering
% \captionsetup{font={small}}
\vspace{-1.0em}
\caption{Workflow of our proposed agents. (a) $\model$ explores the relative knowledge under the user's instruction and provides the reranking results refined by the self-reflection mechanism. (b) $\modelplus$ designs the dynamic memory mechanism to improve the personalized ability of $\model$.}
\label{fig:framework}
\vspace{-1.0em}
\end{figure*}

In this part, we firstly introduce the naive solution $\model$ based on $\dataset$, which can learn the intention from the user instruction. Next, we introduce our $\modelplus$ equipped with individual dynamic memory. The workflow of models are shown in Fig. \ref{fig:framework}. All the prompt templates used in $\model$ and $\modelplus$ and examples of responses are provided in Appendix~\ref{appendix::prompt template}.

\subsection{$\model$}
\noindent \textbf{Parser.} The user's instructions encompass both direct lower-level demands and hidden higher-order preferences. Addressing these higher-order preferences requires agents to be equipped with relevant knowledge, transforming them into domain-specific experts that serve the user. Domain-specific experts use their professional knowledge to recognize differences between products, such as parameterized variations, and connect these distinctions to the user's expressed needs. The parser model is built upon a large language model (LLM), represented by \( M_p \), which is specifically prompted to generate internal knowledge and decide whether to use external tools to extract knowledge from the open world based on the given instruction. In the first step, we concatenate the instruction \( X_I \) with the parser's prompt template \( P_{tp} \) and prompt the LLM to output the related internal knowledge \( X_{IK} \) about the instruction. This step also involves deciding whether to use external tools \( O_T \) and generating the instruction keywords \( X_{KW} \). For example, in the book domain, this may include understanding each book's theme, types of storylines, and other related aspects. Next, if the parser \( M_p \) decides to use external tools, the instruction keywords \( X_{KW} \) and the potential tool options \( O_T \) are utilized to explore the external knowledge \( X_{EK} \). 
% To enhance reasoning and planning capabilities, the React framework~\citep{react} is utilized within the parser model.
% react framework

\begin{small}
\begin{equation}
    O_T,X_{KW},X_{IK} \leftarrow M_p(X_I \, \Vert \, P_{tp}) ; \;  X_{EK} \leftarrow M_p(O_T \, \Vert \, X_{KW})
\end{equation}
\end{small}

\noindent \textbf{Reranker.} After obtaining the instruction-related knowledge, the reranker, denoted by the LLM-based model \( M_r \), reranks the initial ranking list \( \mathcal{R} \) from the recommender platform.
% and generates an explanation. 
In addition to the generated knowledge \( X_{IK} \) and \( X_{EK} \), we incorporate the user's historical sequential information \( X_{SU} \), which serves as a static memory of the user. Similarly, the textual information \( X_{Item} \) of the items in the ranking list is also provided. Overall, the instruction-related knowledge, the textual information \( X_{SU} \) and \( X_{Item} \), along with the reranker's prompt template \( P_{tr} \), are fed into the reranker \( M_r \). Formally, this process can be written as follows:

\vspace{-10pt}
\begin{small}
\begin{equation}
\mathcal{R^*} \leftarrow M_r( X_{IK}  \, \Vert  X_{EK} \, \Vert  X_{SU} \, \Vert \, X_{Item} \,  \Vert \, P_{tr})
\label{method:reranker}
\end{equation}
\end{small}

where \( \mathcal{R^*} \) is the reranked item lists and \( X_{Item} \) includes the textual information (such as title and description) of the candidate items and item index from the initial ranking list \( \mathcal{R} \).

\noindent \textbf{Self-reflection Mechanism.} Large language models output content in a generative manner, which can lead to hallucination problems~\citep{huang2023survey}. To address this, we designed a self-reflection mechanism to verify the content of the re-ranked item list. Specifically, we compare the elements between the reranking list and the previous one. If no differences are found, the results are directly output. However, if discrepancies are detected, the self-reflection module invokes the reranker to regenerate the reranking list, adding a prompt $P_{sr}$ to ensure alignment with the original ranked list. The formulation remains the same as in Eq. \ref{method:reranker}, with the prompt $P_{tr}$ replaced by $P_{sr}$.

\subsection{$\modelplus$}
Although our basic framework $\model$ can explore knowledge based on the user's instructions, it fails to effectively model the dynamic interests within the instructions and cannot learn from user feedback.. 
To address this, we design a profile generator to build user's personal profile that learns from the user feedback and a dynamic extractor to extract dynamic interest and build dynamic profile according to the instruction. Unlike existing recommendation models, $\modelplus$ is uniquely optimized for individual users, remaining unaffected by the behaviors of other users.

\noindent \textbf{Profile Generator.} 
In our profile generator, we simulate the training process of a neural network by first feeding training data pairs into the generator, followed by presenting the ground truth interacted item and the corresponding reviews. 
Consider a user with a sequence of interactions, where the most recent interacted item is selected as the positive sample, and a negative item is randomly selected from the non-interacted items. 
The sampled pair, along with their corresponding textual information, are combined and fed into the generator \( M_{ge} \), which selects one item from the two as the recommended item for the user.
Moreover, the user’s static memory \( X_{SU} \) and the rank prompt template \( P_{pr1} \) are also input into the model. Formally, this process can be expressed as:

\vspace{-7pt}
\begin{small}
\begin{equation}
 X_G^T  \leftarrow   M_{ge}( X_{SU} \, \Vert \, X_i^+ \,  \Vert \, X_i^-\,  \Vert \mathcal{F}^{T-1} \,  \Vert P_{pr1})
\end{equation}
\end{small}

where \( X_i^+ \) and \( X_i^- \) represent the textual information of the positive and negative samples, respectively, and \( \mathcal{F}^{T-1} \) denotes the user's profile in the previous round of interaction. $X_G^{T}$ is the recommended item generated by $M_{ge}$. $T$ represents the round of feedback update iterations.
Then, we incorporate user feedback to further update the user's profile in this round. This feedback includes the groundtruth interacted item and any optional reviews. The generator \( M_{ge} \) integrates this information as follows:

\vspace{-7pt}
\begin{small}
\begin{equation}
    \mathcal{F}^{T} \leftarrow M_{ge}(\mathcal{F}^{T-1} \, \Vert  X_i^{+*} \, \Vert  X_G^{T} \, \Vert P_{pr2} )
\end{equation}
\end{small}

where \( X_i^{+*} \) contains the positive sample's textual information augmented with feedback data, and \( P_{pr2} \) is the corresponding prompt template. 
% For the initialization of a user's profile, we perform the procedure only once, rather than repeating it $T$ times, in order to reduce computational costs.

\noindent \textbf{Dynamic Extractor.} Similar to the attention mechanism~\citep{transformer}, we propose a dynamic extractor to extract instruction-relative information based on the instruction.
We prompt the extractor ($M_e$) to extract the dynamic interest from the static memory of user historical information $X_{SU}$ and the generated profile $\mathcal{F}_{T}$ according to the instruction $X_I$ and the generated instruction-related knowledge $X_{IK}$ and $X_{EK}$.
% more about advantages.
It can be formulated as: 

\vspace{-7pt}
\begin{small}
\begin{equation}
  \mathcal{F}_{d}^{T} \, ,X_{DU} \leftarrow M_{e}(\mathcal{F}^{T} \, \Vert  X_{SU} \, \Vert X_I  \, \Vert X_{IK}  \, \Vert X_{EK} \, \Vert P_{e}  )
\end{equation}
\end{small}

where $\mathcal{F}_{d}^{T}$ and $X_{DU}$ represents the dynamic profile and dynamic interest, respectively. These two components form the dynamic memory. $P_{e}$ is the prompt template. 

\noindent \textbf{Reranker.} After constructing the dynamic memory of a user, the reranker utilizes the information to generate the reranked results. Similar to Eq. \ref{method:reranker}, it can be expressed as:

\vspace{-7pt}
\begin{small}
\begin{equation}
\mathcal{R^*} \leftarrow M_r( X_{IK}  \, \Vert  X_{EK} \, \Vert  X_{SU}  \, \Vert \, \mathcal{F}_{d}^{T}  \, \Vert \, X_{DU} \, \Vert \, X_{Item} \,  \Vert \, P_{tr}^*)
\label{method:reranker2}
\end{equation}
\end{small}

where \( P_{tr}^* \) represents the prompt template for the reranker in \( \modelplus \). Besides, a self-reflection mechanism is also implemented to ensure consistent results, using the same inputs as the reranker, except for the prompt template.

\section{Empirical Evaluation}
In this section, we present extensive experiments to demonstrate the effectiveness of $\model$ and $\modelplus$, aiming to answer the following four research questions (\textbf{RQs}).

\noindent $\bullet$ \textbf{RQ1}: How does the performance of $\model$ and $\modelplus$ compare to state-of-the-art baselines across various datasets?

\noindent $\bullet$ \textbf{RQ2}: Can our method mitigate the echo chamber effect by helping users filter out unwanted ads and recommending more diverse items, rather than just recommending popular ones?

\noindent $\bullet$ \textbf{RQ3}: How well does our method perform for both active and less-active user groups?

\noindent $\bullet$ \textbf{RQ4}: Are the proposed reranker and self-reflection mechanism effective in practice?

\subsection{Experiment Setup} 

\noindent \textbf{Dataset.} Given the absence of a recommendation dataset that includes proactive user instructions in the user-agent-platform paradigm, we construct $\dataset$ datasets using existing recommendation datasets, including Amazon~\citep{amazon18dataset}, Yelp\footnote{\url{https://www.kaggle.com/datasets/yelp-dataset/yelp-dataset/versions}}, and Goodreads~\citep{goodreads}. These datasets provide textual information such as item titles, descriptions, and reviews. We eliminate users and items that have fewer than 5 associated actions to ensure sufficient data density. 
For each interaction, we generate the instruction for this interaction based on the corresponding user review and filter through a post-processing verification mechanism. To further enhance the linguistic diversity of the instructions, we assign a persona to each user. More details are in the following.

\noindent \textit{Instruction Generator}: Initially, we manually annotate several instruction-review pairs, providing few-shot examples for LLMs to facilitate in-context learning. These few-shot examples, along with reviews paired with a random persona from Persona Hub~\citep{personahub}, are then fed into the LLM\footnote{We use GPT-4o-mini for data generation.} to generate instructions. To ensure that the few-shot examples remain dynamic, we create a list to store the instruction-review pairs and allow the LLM to decide whether a newly generated instruction should be included as an example. Examples of the annotated instruction-review pairs, generated instructions, and the data construction processes can be found in Appendix~\ref{appendix::dataset}.
% Appendix~\ref{appendix::examplesconstruct}.

\noindent \textit{Instruction Cleaner}: To prevent data leakage from the reviews, we test if or not the LLM can recover the item from the generated instruction. More specifically, given the instruction, we employ the LLM to choose between the ground-truth item and a randomly selected negative item. The LLM generates a certainty score based on the instruction and the item's textual information. Based on the result, we retain all of those instructions for which the LLM cannot infer the ground-truth item, and also keep an equal number of correctly inferred instructions that has low certainty scores.
Statistical analysis of \( \dataset \) dataset is in Table \ref{dataset-stats}. For the filtered instructions and the retained instructions, we 
show some examples in Appendix~\ref{appendix:filter}.
\begin{figure*}[h!]
\centering{
\includegraphics[width=0.80\textwidth]{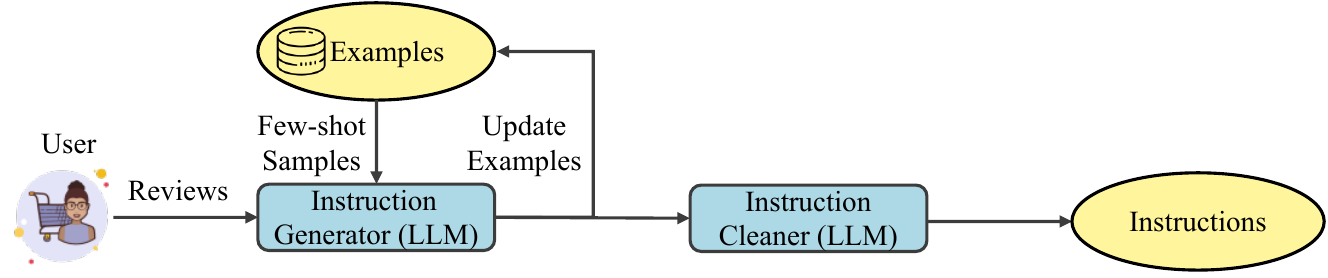}}
% \captionsetup{font={footnotesize}}
% \vspace{-5pt}
\caption{
The overview of our $\dataset$ dataset construction.
}
\vspace{-1.0em}
\label{fig:dataset construction}
\end{figure*}

\begin{table*}
\centering
\caption{Statistics of the $\dataset$ dataset: $\left|\mathcal{U}\right|$, $\left|\mathcal{V}\right|$, and $\left|\mathcal{E}\right|$ represent the number of users, items, and interactions, respectively. \#$\left|X_I\right|$ denotes the average token length of user instructions, while \#$\left|S_U\right|$ represents the average token length of the user's static memory.}
\vspace{-1.0em}
\label{dataset-stats}
% \vspace{-10pt}
% \setlength\tabcolsep {0.3pt}
\resizebox{0.85\textwidth}{!}{
\begin{tabular}{c|cccc|cc}
\midrule
Dataset & $\left|\mathcal{U}\right|$ & $\left|\mathcal{V}\right|$ & $\left|\mathcal{E}\right|$ & Density & \#$\left|X_I\right|$ & \#$\left|S_U\right|$  \\ \midrule
$\dataset$ - Amazon Book       & 7,377                 & 120,925                  & 207,759   & 0.023\%   & 164   & 1276       \\ 
$\dataset$ -Amazon Movietv       & 5,649                & 28,987                  & 79,737    & 0.049\%  & 40     & 726       \\ 
$\dataset$ - Goodreads       &  11,734               & 57,364                   & 618,330  & 0.092\%   & 41      & 2827      \\ 
$\dataset$ - Yelp       & 2,950              & 31,636                   & 63,142    & 0.068\%    & 40      & 1976  \\ \midrule
\end{tabular}}
% \begin{tablenotes}    
%         % \small            
%         \item $\left|\mathcal{U}\right|$, $\left|\mathcal{V}\right|$, $\left|\mathcal{E}\right|$ denote the number of user, item and ratings, respectively. \#$\left|X_I\right|$ is the average token length of user's instruction. \#$\left|S_U\right|$ represents the average token length of user's static memory. 
%       \end{tablenotes}
\vspace{-10pt}
      \end{table*}

\vspace{-7pt} \noindent \textbf{Evaluation Protocol.} We randomly sample 9 negative items with one true item to make the candidate ranking list. 
Following the data split in sequential recommendation~\citep{sasrec}, the most recent interaction is reserved for testing.
% , the second most recent for validation, and all preceding interactions are used for training. 
The agent-based works, including ours, utilize all the interaction data except the most recent one to construct the agent's memory. 
For evaluation metric, we adopt the typical top-$N$ metrics hit rate (HR@\{1, 3\}), normalized discounted cumulative gain (NDCG@\{3\})~\citep{NDCG} and Mean Reciprocal Rank (MRR)~\citep{MRR}. 
In addition to conventional ranking metrics, we conduct additional experiments to ensure that our $\model$/$\modelplus$ can act as a shield between users and the recommendation system. Specifically, we design evaluation metrics such as the percentage of filtered Ads items (FR@{1,3,5,10}) and popularity-weighted ranking metrics (P-HR@{3} and P-MRR) to validate the mitigation of the echo chamber effect~\citep{echochamber,echochamber2}. We use freq$_i$ to denote the frequency of item $i$ in the dataset. Formally, these metrics are defined as:

\begin{small}
\begin{equation}
    % \text{FR@k} = \begin{cases}
    %                 1, & \text{if } r_{ads} > k, \\
    %                 0, & \text{if } r_{ads} \leq k.
    %                 \end{cases} \;\;\;\;\;\;\;\;  \text{P-Rank} = \left(1 - \sigma\left(\lambda \cdot \text{freq}_i\right)\right) \cdot \text{Rank}.
    \text{FR@k} = \begin{cases}
                    1, & \text{if } r_{Ads} > k, \\
                    0, & \text{if } r_{Ads} \leq k.
                    \end{cases}
\end{equation}
\end{small}

\begin{small}
    \begin{equation}
        \text{P-Rank} = \left(1 - \sigma\left( \text{freq}_i\right)\right) \cdot \text{Rank}.
    \end{equation}
\end{small}
where $r_{Ads}$ denotes the position of Ads items in the re-ranked list, \text{Rank} represents ranking metrics such as HR, and $\sigma$ refers to the sigmoid function. The Ads items is randomly selected from a different data domain. For example, to simulate the Ads items in $\dataset$ - Amazon Book, we select Ads items from the data in $\dataset$ - Amazon Movietv, to test if the agent is able to demote an irrelevant item even if the item is already added into the ranking list by the recommender system.
% while $\lambda$ serves as a scaling factor. 
Additionally, we report the performance for both active and less-active users separately~\citep{fairnessrec}. We also analyze the probability of changes in the top-ranked items after reranking. To further assess the effectiveness of our self-reflection mechanism, we report the occurrence rate of hallucination. For all evaluation metrics in our experiments, higher values indicate better performance.

\noindent \textbf{Baselines.} We compare our method with three classes of baselines: (1) Sequential recommendation methods, i.e.,  BERT4Rec~\citep{bert4rec}, GRU4Rec~\citep{gru4rec} and SASRec~\citep{sasrec}. (2) Instruction-aware methods, i.e., BM25~\citep{bm25}, BGE-Rerank~\citep{bge_embedding} and EasyRec~\citep{ren2024easyrec}. (3) Recommendation agents, i.e., ToolRec~\citep{toolrec} and AgentCF~\citep{agentcf}. 
% For the sequential recommendation baselines, only item ID information is considered in the model. In instruction-aware methods, the user's instruction is treated either as a query or used to build the user's profile. For a fair comparison and to optimize computational efficiency, the number of memory-building rounds in AgentCF is set to 1, the same as in our $\modelplus$. 
Detailed implementation and introduction of baselines are in Appendix~\ref{appendix::experiment results}.
 
\subsection{Performance Comparison}

\noindent \textbf{Main Results. (RQ1)} Tables~\ref{tab:exper-result1} and \ref{tab:exper-result2} present the experimental results across four datasets using different evaluation metrics. By incorporating instruction knowledge into the model, the instruction-aware baselines outperform traditional recommendation agent methods. Benefiting from the alignment with collaborative filtering and natural language information, EasyRec pretraining on several Amazon datasets achieves the second-best results, trailing only our $\model$. Our $\modelplus$ outperforms the second-best baseline, EasyRec, with the averagely 16.6\% improvement. This improvement is partly attributed to the parser component, which learns instruction-aware knowledge, enabling the reranker to better understand the user's intentions. Meanwhile, our proposed dynamic memory component leverages user feedback to construct a more accurate user profile and dynamically extract interests from historical data based on the instruction.

\begin{table*}[tb!]
\small
		\centering
		\caption{Evaluation results (\%) of the ranking metric ($\uparrow$) on the $\dataset$. We highlight the methods with the \textbf{\textcolor{teal}{first}}, \textbf{\textcolor{purple}{second}} and \textbf{third} best performances. \label{tab:exper-result1}}
  \vspace{-1.0em}
  % \vspace{-10pt}
		
		\resizebox{0.95\textwidth}{!}{
		\begin{tabular}{c|cccc|cccc}
		\midrule
		\multirow{2}{*}{\textbf{Model}}  & \multicolumn{4}{c|}{\texttt{$\dataset$ - Amazon Book}} & \multicolumn{4}{c}{\texttt{$\dataset$ - Amazon Movietv}} \\
         & \textbf{HR@1} & \textbf{HR@3} & \textbf{NDCG@3} & \textbf{MRR} & \textbf{HR@1} & \textbf{HR@3} & \textbf{NDCG@3} & \textbf{MRR} \\
		\midrule
		GRU4Rec  & 11.00 & 31.41 & 22.53 & 30.10 & 15.80 & 36.85 & 27.63 & 34.36 \\

            BERT4Rec  & 11.48 & 30.90 & 22.32 & 30.31 & 14.74 & 35.13 & 26.36 & 33.43 \\
            
            SASRec  &  11.08 & 31.34 & 22.42 & 30.15 & 34.52 & 49.71 & 43.18 & 48.06\\

             \midrule

            BM25  & 9.92 & 24.48 & 18.21 & 27.00 & 11.29 & 30.27 & 22.09 & 30.04 \\
            
            BGE-Rerank  & 25.36 & 45.90 & 37.11 & 42.84 & 25.44 & 47.48 & 38.02 & 43.28\\
            
            EasyRec  & \textbf{30.70} & \textbf{48.87} & \textbf{41.09} & \textbf{46.14} & \textbf{34.96} & \color{purple}\textbf{61.30} & \color{purple}\textbf{50.15} & \textbf{52.98} \\
            
             \midrule
            ToolRec & 10.56 & 30.60 & 21.88 & 29.77 & 13.84 & 35.67 & 26.20 & 33.21 \\

            AgentCF  & 14.24 & 34.16 & 25.55 & 32.77 & 25.90 & 49.82 & 39.64 & 44.23 \\
            
            \midrule
            
            {\model}  & \color{purple}\textbf{31.89} & \color{purple}\textbf{48.99} & \color{purple}\textbf{41.69} & \color{purple}\textbf{47.23} & \color{purple}\textbf{38.19} & \textbf{56.87} & \textbf{48.93} & \color{purple}\textbf{53.04} \\

            {\modelplus}  & \color{teal}\textbf{35.11} & \color{teal}\textbf{53.51} & \color{teal}\textbf{45.64} & \color{teal}\textbf{50.28} &  \color{teal}\textbf{46.43} & \color{teal}\textbf{65.77} & \color{teal}\textbf{57.67} & \color{teal}\textbf{60.43} \\

  %           \midrule
		% {\modelplus} & Yes & \color{teal}\textbf{34.35} & \color{teal}\textbf{52.91} & \color{teal}\textbf{45.04} & 48.61 &  \color{teal}\textbf{46.01} & \color{teal}\textbf{65.97} & \color{teal}\textbf{57.58} & 60.17 \\

		\midrule
	\end{tabular}				
		}
    \vspace{-10pt}
\end{table*}

\begin{table*}[tb!]
		\centering
		\caption{Evaluation results (\%) of the ranking metric ($\uparrow$) on $\dataset$.}
  \vspace{-1.0em}
  % \vspace{-10pt}
		\label{tab:exper-result2}
		\resizebox{0.95\textwidth}{!}{
		\begin{tabular}{c|cccc|cccc}
		\midrule
		\multirow{2}{*}{\textbf{Model}}  & \multicolumn{4}{c|}{\texttt{$\dataset$ - Goodreads}} & \multicolumn{4}{c}{\texttt{$\dataset$ - Yelp}} \\
       & \textbf{HR@1} & \textbf{HR@3} & \textbf{NDCG@3} & \textbf{MRR} & \textbf{HR@1} & \textbf{HR@3} & \textbf{NDCG@3} & \textbf{MRR} \\
		\midrule
		GRU4Rec  & 15.36 & 39.52 & 29.08 & 35.41 & 10.94 & 30.67 & 21.88 &29.70 \\

            BERT4Rec  &12.70 & 34.69 & 25.02 & 32.32 & 10.99 & 31.02 & 22.32 & 30.05 \\
 
            SASRec  &  18.52 & 41.24 & 31.47 & 37.60 & 12.59 & 31.09 & 22.65 & 30.15\\

            \midrule

            BM25  & 14.25 & 40.34 & 29.01 & 35.40 & 12.85 & 33.08 & 24.34 & 31.85 \\

            BGE-Rerank & 17.26 & 40.82 & 30.60 & 36.97  & \textbf{33.05} & 55.29 & 45.70 & \textbf{49.90}\\
            
            EasyRec  & 13.94 & 35.38 & 26.11 & 33.27  & 32.41 & \textbf{56.31} & \textbf{46.04} & 49.86 \\
            
            \midrule
            ToolRec & 19.06 & 42.79 & 32.61 & 38.44 & 12.07 & 30.92 & 22.83 & 30.21 \\

            AgentCF  & \textbf{21.61} & \textbf{46.09} & \textbf{35.60} & \textbf{40.96} & 13.36 & 34.83 & 25.66 & 32.61 \\

            \midrule
            
            {\model}  & \color{purple}\textbf{23.56} & \color{purple}\textbf{47.01} & \color{purple}\textbf{36.98} & \color{purple}\textbf{42.19} & \color{purple}\textbf{37.40} & \color{purple}\textbf{56.33} & \color{purple}\textbf{48.28} & \color{purple}\textbf{52.42} \\

            {\modelplus }  & \color{teal}\textbf{30.97} & \color{teal}\textbf{56.69} & \color{teal}\textbf{45.76} & \color{teal}\textbf{49.14} &  \color{teal}\textbf{39.22} & \color{teal}\textbf{57.92} & \color{teal}\textbf{49.96} & \color{teal}\textbf{53.78} \\

  %           \midrule
		% {\modelplus} & Yes & \color{teal}\textbf{34.35} & \color{teal}\textbf{52.91} & \color{teal}\textbf{45.04} & 48.61 &  \color{teal}\textbf{46.01} & \color{teal}\textbf{65.97} & \color{teal}\textbf{57.58} & 60.17 \\

		\midrule
	\end{tabular}				
		}
    \vspace{-5pt}
\end{table*}

\noindent \textbf{Echo Chamber Effect. (RQ2)} We also report the experimental results evaluating the echo chamber effect in Table~\ref{tab:echo-result1}. Ads items are randomly inserted into the candidate ranking list from other domains to simulate advertising scenarios that users may have encountered. To mitigate position bias in LLMs~\citep{llmbias}, Ads items are added randomly within the candidate list positions. $\modelplus$ accurately identifies users' instructions and extracts knowledge about their underlying needs, thereby effectively removing undesired Ads. Benefitting from not being trained in a purely data-driven manner and constructing user profiles based on their feedback, our $\modelplus$ also recommends more diverse items to users (both active and less-active items), instead of focusing solely on popular items, and meanwhile improves the overall recommendation performance. Drawing from these experimental results, we conclude that our $\modelplus$ can mitigate the echo chamber effect and act as a protective shield for users.
Due to the page limitation, we provide full experiment results in Appendix~\ref{appendix:echo chamber}.

\begin{table*}[tb!]
		\centering
		\caption{Evaluation of the echo chamber effects (\%) ($\uparrow$) on $\dataset$. }
  \vspace{-1.0em}
  % \vspace{-10pt}
		\label{tab:echo-result1}
		\resizebox{0.95\textwidth}{!}{
		\begin{tabular}{c|cccc|cccc}
		\midrule
		\multirow{2}{*}{\textbf{Model}}  & \multicolumn{4}{c|}{\texttt{$\dataset$ - Amazon Book}} & \multicolumn{4}{c}{\texttt{$\dataset$ - Yelp}} \\
         & \textbf{FR@1} & \textbf{FR@3}   & \textbf{P-HR@3} & \textbf{P-MRR} & \textbf{FR@1} & \textbf{FR@3}   & \textbf{P-HR@3} & \textbf{P-MRR} \\
		\midrule
            
            EasyRec  & \textbf{68.41} & \textbf{64.32} & \textbf{59.28} & \textbf{56.09} & \textbf{76.45} & \textbf{66.50} & \textbf{61.05} & \textbf{56.85} \\

            ToolRec & 70.13 & 66.61 & 36.74 & 35.80 & 72.64 & 63.64 & 32.50 & 32.73 \\

            AgentCF  & 58.02 & 50.04 & 41.10 & 39.42 & 71.30 & 64.15 & 38.46 & 36.44 \\
            
            \midrule
            
            {\model}  & \color{purple}\textbf{71.98} & \color{purple}\textbf{67.82} & \color{purple}\textbf{59.51} & \color{purple}\textbf{57.32} & \color{purple}\textbf{78.24} & \color{purple}\textbf{69.71} & \color{purple}\textbf{62.74} & \color{purple}\textbf{58.76} \\

            {\modelplus}  & \color{teal}\textbf{77.15} & \color{teal}\textbf{70.15} & \color{teal}\textbf{64.70} & \color{teal}\textbf{60.87} &  \color{teal}\textbf{87.69} & \color{teal}\textbf{84.20} & \color{teal}\textbf{64.48} & \color{teal}\textbf{60.20} \\

		\midrule
	\end{tabular}				
		}
    \vspace{-5pt}
\end{table*}

\noindent \textbf{Protect Less-Active Users. (RQ3)} We define the top 20\% of users as active, with the remaining 80\% classified as less-active~\citep{fairnessrec,nmcdrhead}. Since our data is sampled and filtered using a 10-core process, most users exhibit rich behavioral patterns. Consequently, active users tend to show poorer performance compared to less-active users, largely due to the decline in LLM performance with longer texts~\citep{liu2024lost}.
As illustrated in Table~\ref{tab:active-result}, our $\modelplus$ enhances the performance for both active and less-active users. For less-active users, we construct individual profiles based on their feedback, ensuring that these profiles are not influenced by other users. The experimental results demonstrate that our dynamic memory mechanism offers personalized services tailored to each user individually. Detailed implementation and introduction of baselines are in Appendix~\ref{appendix:active}.

\begin{table*}[tb!]
		\centering
		\caption{The performance (\%) of active  and less-active users on  $\dataset$ - Amazon book. }
  \vspace{-1.0em}
  % \vspace{-10pt}
		\label{tab:active-result}
		\resizebox{0.95\textwidth}{!}{
		\begin{tabular}{c|cccc|cccc}
		\midrule
		\multirow{2}{*}{\textbf{Model}}  & \multicolumn{4}{c|}{\texttt{Less-Active Users}} & \multicolumn{4}{c}{\texttt{Active Users}} \\
         & \textbf{HR@1} & \textbf{HR@3} & \textbf{NDCG@3} & \textbf{MRR} & \textbf{HR@1} & \textbf{HR@3} & \textbf{NDCG@3} & \textbf{MRR} \\
		\midrule
            
            EasyRec & \textbf{32.93}	 & \textbf{51.07}	& \textbf{43.32} &	\textbf{48.04}	& \textbf{28.71}	&\textbf{47.64}	& \textbf{39.53}	& \textbf{44.61} \\

            ToolRec & 10.57 &	30.86	& 22.01	& 29.88	& 10.04	& 31.73	& 22.32	& 29.54 \\

            AgentCF  & 14.79	& 35.00	 & 26.26	& 33.35	& 14.87	& 34.37	& 25.93	& 33.24 \\
            
            \midrule
            
            {\model}  & \color{purple}\textbf{34.07} & \color{purple}\textbf{50.79} & \color{purple}\textbf{43.67} & \color{purple}\textbf{49.00} & \color{purple}\textbf{29.96} & \color{purple}\textbf{47.73} & \color{purple}\textbf{40.14} & \color{purple}\textbf{45.71} \\

            {\modelplus}  & \color{teal}\textbf{37.92} & \color{teal}\textbf{55.75} & \color{teal}\textbf{47.84} & \color{teal}\textbf{52.11} &  \color{teal}\textbf{33.27} & \color{teal}\textbf{51.74} & \color{teal}\textbf{43.81} & \color{teal}\textbf{48.67} \\

		\midrule
	\end{tabular}				
		}
    \vspace{-5pt}
\end{table*}

\begin{figure}[tb!]
\centering
% 第一行的两个子图
\subfigure[$\dataset$-Amazon books]
{
\begin{minipage}[t]{0.32\linewidth}
\centering
\includegraphics[width=0.97\linewidth]{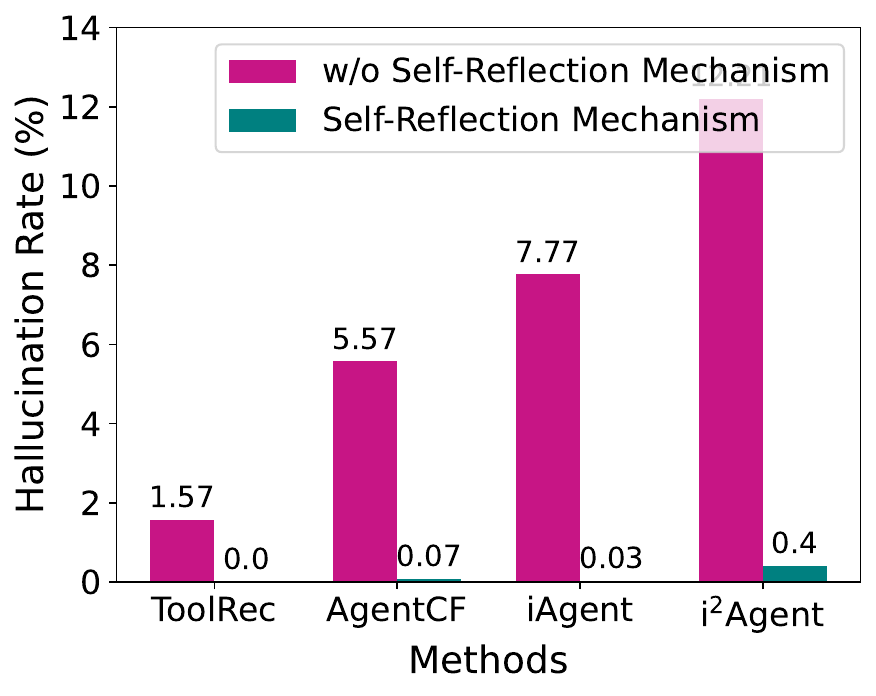}
%\caption{fig1}
\end{minipage}%
}%
\hfill % 填充所有可用的水平空间
\subfigure[$\dataset$-Goodreads]
{
\begin{minipage}[t]{0.32\linewidth}
\centering
\includegraphics[width=0.97\linewidth]{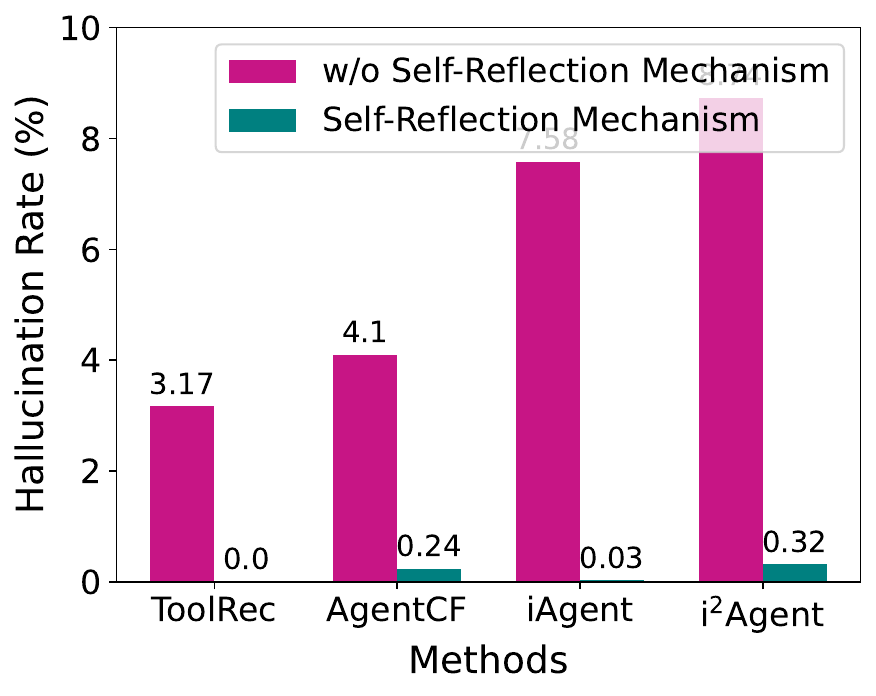}
%\caption{fig2}
\end{minipage}%
}% 
\hfill
\subfigure[$\dataset$-Yelp]
{
\begin{minipage}[t]{0.32\linewidth}
\centering
\includegraphics[width=0.97\linewidth]{figure/amazonbooks_sr.pdf}
%\caption{fig2}
\end{minipage}%
}%
\vspace{-5pt}
\\
% 第二行的两个子图
\subfigure[$\dataset$-Amazon books]{
\begin{minipage}[t]{0.32\linewidth}
\centering
\includegraphics[width=0.97\linewidth]{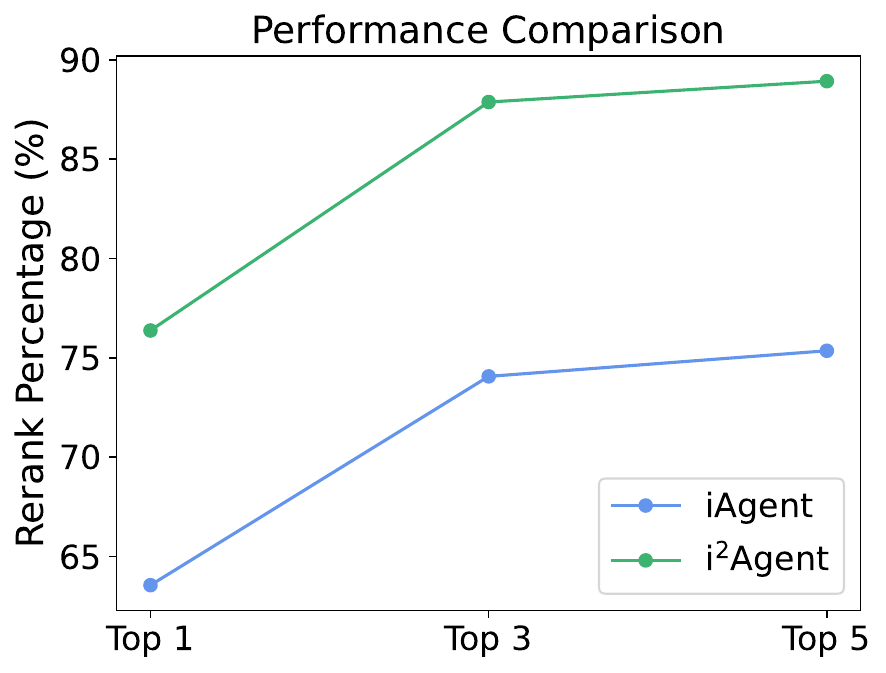}
%\caption{fig1}
\end{minipage}%
}%
\hfill % 填充所有可用的水平空间
\subfigure[$\dataset$-Goodreads]{
\begin{minipage}[t]{0.32\linewidth}
\centering
\includegraphics[width=0.97\linewidth]{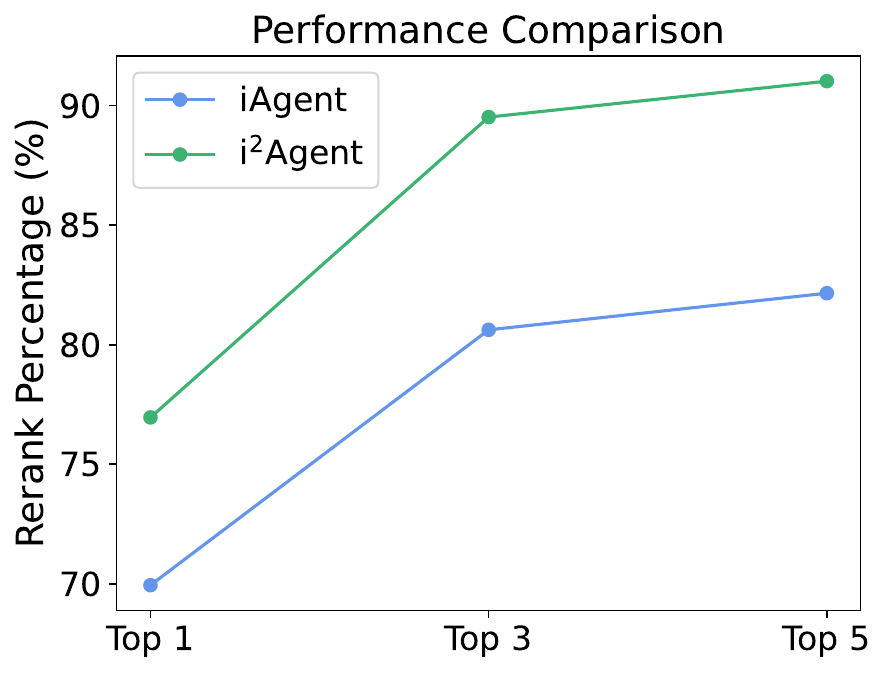}
%\caption{fig2}
\end{minipage}%
}% 
\hfill
\subfigure[$\dataset$-Yelp]{
\begin{minipage}[t]{0.32\linewidth}
\centering
\includegraphics[width=0.97\linewidth]{figure/amazonbooks_rerank.pdf}
%\caption{fig2}
\end{minipage}%
}%
\centering
% \captionsetup{font={small}}
\vspace{-8pt}
\caption{The first row presents the hallucination rate with and without the self-reflection mechanism, while the second row illustrates the probability of changes in the ranking list after our reranker.}
\vspace{-15pt}
\label{fig:model study}
\end{figure}

\noindent \textbf{Model Study. (RQ4)} First, we analyze the impact of our self-reflection mechanism on the LLM's hallucination rate. When implementing ToolRec~\citep{toolrec} and AgentCF~\citep{agentcf}, we applied the self-reflection mechanism to improve the accuracy of the reranking list. As shown in Fig.~\ref{fig:model study}, the self-reflection mechanism reduces the hallucination rate by at least 20-fold. In this mechanism, we prompt the LLM to generate the reranking list based on the initial ranking list. However, $\modelplus$ exhibits the highest error rate, as the longer text sequence causes the LLM to lose some information from the original ranking list. Based on the experimental results, we can safely conclude that our self-reflection mechanism effectively alleviates LLM-induced hallucinations.

Next, we examine the re-ranking ratio across our models. We compare whether the elements in the ranking list change before and after reranking, focusing on the top@\{1,3,5\} positions. If any element changes position, it is considered a rerank. The results indicate that changes occur almost every time during reranking, suggesting that our agent is consistently performing personalized reranking on the list generated by the recommender platform.

\section{Related Work}
\subsection{Recommender System}
Sequential recommendation models~\citep{gru4rec} focus on developing temporal encoders to capture both short- and long-term user interests, exemplified by SASRec~\citep{sasrec}'s attention mechanism and BERT4Rec~\citep{bert4rec}'s bidirectional encoder. Recent developments have integrated large language models, with some approaches treating item indices as tokens for generative recommendations~\citep{p5}, while others utilize LLMs~\citep{e4srec,slmrec} as sequential embedding extractors to enhance recommendation performance. The rise of LLMs has also transformed conversational recommendation systems (CRS)\citep{2018CRS,zhang2018CRS}, improving dialogue understanding and flexibility\citep{LLM+CRSgoogle,LLM+CRS2} compared to conventional approaches that were limited in dialogue format and turn count.

\subsection{Personal Language-based Agent}
With the advancement of large language models~\citep{gpt4}, research has evolved from early persona-based dialogue agents~\citep{zhang2018personalizing, park2023generative} to more sophisticated domain-specific agents~\citep{webagent,mind2web,travelagent} incorporating tool learning and memory mechanisms. In the recommendation domain, recent work has developed recommendation agents \citep{toolrec,wang2023recmind,generativerecagent} that leverage historical interaction information as user memory and utilize LLMs for ranking. Notably, newer approaches focus on user-side operations \citep{wang2024macrec,huang2023recommender}, generating re-ranking results based on user instructions and individual memory. We provide a more detailed related work in Appendix~\ref{sec:appendix:related work}.
 
\section{Conclusion}
% \section{Actionable Insights}
In this work, we first design a straightforward instruction-aware agent ($\model$) to analyze user instructions and integrate relevant and comprehensive knowledge. Moreover, to enhance the agent's personalized abilities, we propose individual instruction-aware agent ($\modelplus$), which incorporates a dynamic memory mechanism to learn from user's personal feedback and extracts the dynamic interests. In addition to these technical contributions, our work also presents unique and complementary avenues for future research. We discuss potential future directions and open challenges in Appendix~\ref{sec:future work}.

\section{Limitation}
While our work demonstrates promising results, there are several limitations to note. First, our current implementation primarily focuses on English instructions, and the effectiveness of the model across different languages remains to be explored. recommendation scenarios. Additionally, while our evaluation metrics show improvements in recommendation quality, they may not fully capture the nuanced aspects of user satisfaction and long-term engagement. These limitations suggest potential directions for future research in developing more efficient and comprehensive instruction-aware recommendation systems.
% 1. [improve performance] trainable reranker or utilize existing ranking model as tools but combine with existing personalized memory mechanism.

% 2. [fair comparison] long-text more negative samples fair comparison. 

% 3. [More Interpretable]

% 4. [Agents interact with RecSys / autonomus interaction/ mutual learning ]

% 1. Rank as tools / 

% 2. Multimodal information input

% 3. Long-context more negative samples.

% 4. User agent and RecSys mutual learning/improvement

% 5. Next generation rec promote trustworthy recsys, more interpretable (text saved memory – dynamic process accessible.), fairness (user instruction guided), | connect other task such as social network

\bibliography{acl}

\begin{thebibliography}{72}
\providecommand{\natexlab}[1]{#1}

\bibitem[{Abdin et~al.(2024)Abdin, Jacobs, Awan, Aneja, Awadallah, Awadalla, Bach, Bahree, Bakhtiari, Behl et~al.}]{phi3.8b}
Marah Abdin, Sam~Ade Jacobs, Ammar~Ahmad Awan, Jyoti Aneja, Ahmed Awadallah, Hany Awadalla, Nguyen Bach, Amit Bahree, Arash Bakhtiari, Harkirat Behl, et~al. 2024.
\newblock Phi-3 technical report: A highly capable language model locally on your phone.
\newblock \emph{arXiv preprint arXiv:2404.14219}.

\bibitem[{Achiam et~al.(2023)Achiam, Adler, Agarwal, Ahmad, Akkaya, Aleman, Almeida, Altenschmidt, Altman, Anadkat et~al.}]{gpt4}
Josh Achiam, Steven Adler, Sandhini Agarwal, Lama Ahmad, Ilge Akkaya, Florencia~Leoni Aleman, Diogo Almeida, Janko Altenschmidt, Sam Altman, Shyamal Anadkat, et~al. 2023.
\newblock Gpt-4 technical report.
\newblock \emph{arXiv preprint arXiv:2303.08774}.

\bibitem[{Afsar et~al.(2022)Afsar, Crump, and Far}]{rl-rec}
M~Mehdi Afsar, Trafford Crump, and Behrouz Far. 2022.
\newblock Reinforcement learning based recommender systems: A survey.
\newblock \emph{ACM Computing Surveys}, 55(7):1--38.

\bibitem[{Aguirre et~al.(2015)Aguirre, Mahr, Grewal, De~Ruyter, and Wetzels}]{nopersonalized}
Elizabeth Aguirre, Dominik Mahr, Dhruv Grewal, Ko~De~Ruyter, and Martin Wetzels. 2015.
\newblock Unraveling the personalization paradox: The effect of information collection and trust-building strategies on online advertisement effectiveness.
\newblock \emph{Journal of retailing}, 91(1):34--49.

\bibitem[{Aslay et~al.(2018)Aslay, Matakos, Galbrun, and Gionis}]{polar1}
Cigdem Aslay, Antonis Matakos, Esther Galbrun, and Aristides Gionis. 2018.
\newblock Maximizing the diversity of exposure in a social network.
\newblock In \emph{2018 IEEE international conference on data mining (ICDM)}, pages 863--868. IEEE.

\bibitem[{Bakshy et~al.(2015)Bakshy, Messing, and Adamic}]{echoeffect1}
Eytan Bakshy, Solomon Messing, and Lada~A Adamic. 2015.
\newblock Exposure to ideologically diverse news and opinion on facebook.
\newblock \emph{Science}, 348(6239):1130--1132.

\bibitem[{Chan et~al.(2024)Chan, Wang, Yu, Mi, and Yu}]{personahub}
Xin Chan, Xiaoyang Wang, Dian Yu, Haitao Mi, and Dong Yu. 2024.
\newblock Scaling synthetic data creation with 1,000,000,000 personas.
\newblock \emph{arXiv preprint arXiv:2406.20094}.

\bibitem[{Chaney et~al.(2018)Chaney, Stewart, and Engelhardt}]{chamberother1}
Allison~JB Chaney, Brandon~M Stewart, and Barbara~E Engelhardt. 2018.
\newblock How algorithmic confounding in recommendation systems increases homogeneity and decreases utility.
\newblock In \emph{Proceedings of the 12th ACM conference on recommender systems}, pages 224--232.

\bibitem[{Cheng et~al.(2016)Cheng, Koc, Harmsen, Shaked, Chandra, Aradhye, Anderson, Corrado, Chai, Ispir et~al.}]{wideanddeep}
Heng-Tze Cheng, Levent Koc, Jeremiah Harmsen, Tal Shaked, Tushar Chandra, Hrishi Aradhye, Glen Anderson, Greg Corrado, Wei Chai, Mustafa Ispir, et~al. 2016.
\newblock Wide \& deep learning for recommender systems.
\newblock In \emph{Proceedings of the 1st workshop on deep learning for recommender systems}, pages 7--10.

\bibitem[{Chitra and Musco(2020)}]{echoeffect2}
Uthsav Chitra and Christopher Musco. 2020.
\newblock Analyzing the impact of filter bubbles on social network polarization.
\newblock In \emph{Proceedings of the 13th International Conference on Web Search and Data Mining}, pages 115--123.

\bibitem[{Deng et~al.(2024)Deng, Gu, Zheng, Chen, Stevens, Wang, Sun, and Su}]{mind2web}
Xiang Deng, Yu~Gu, Boyuan Zheng, Shijie Chen, Sam Stevens, Boshi Wang, Huan Sun, and Yu~Su. 2024.
\newblock Mind2web: Towards a generalist agent for the web.
\newblock \emph{Advances in Neural Information Processing Systems}, 36.

\bibitem[{Edizel et~al.(2020)Edizel, Bonchi, Hajian, Panisson, and Tassa}]{edizel2020fairecsys}
Bora Edizel, Francesco Bonchi, Sara Hajian, Andr{\'e} Panisson, and Tamir Tassa. 2020.
\newblock Fairecsys: mitigating algorithmic bias in recommender systems.
\newblock \emph{International Journal of Data Science and Analytics}, 9:197--213.

\bibitem[{Feng et~al.(2023)Feng, Liu, Xue, Cai, Hu, Jiang, Gai, and Sun}]{LLM+CRS2}
Yue Feng, Shuchang Liu, Zhenghai Xue, Qingpeng Cai, Lantao Hu, Peng Jiang, Kun Gai, and Fei Sun. 2023.
\newblock A large language model enhanced conversational recommender system.
\newblock \emph{arXiv preprint arXiv:2308.06212}.

\bibitem[{Friedman et~al.(2023)Friedman, Ahuja, Allen, Tan, Sidahmed, Long, Xie, Schubiner, Patel, Lara et~al.}]{LLM+CRSgoogle}
Luke Friedman, Sameer Ahuja, David Allen, Zhenning Tan, Hakim Sidahmed, Changbo Long, Jun Xie, Gabriel Schubiner, Ajay Patel, Harsh Lara, et~al. 2023.
\newblock Leveraging large language models in conversational recommender systems.
\newblock \emph{arXiv preprint arXiv:2305.07961}.

\bibitem[{Gao et~al.(2021)Gao, Lei, He, de~Rijke, and Chua}]{CRSsurvey}
Chongming Gao, Wenqiang Lei, Xiangnan He, Maarten de~Rijke, and Tat-Seng Chua. 2021.
\newblock Advances and challenges in conversational recommender systems: A survey.
\newblock \emph{AI open}, 2:100--126.

\bibitem[{Garimella et~al.(2018)Garimella, De~Francisci~Morales, Gionis, and Mathioudakis}]{echoeffect3}
Kiran Garimella, Gianmarco De~Francisci~Morales, Aristides Gionis, and Michael Mathioudakis. 2018.
\newblock Political discourse on social media: Echo chambers, gatekeepers, and the price of bipartisanship.
\newblock In \emph{Proceedings of the 2018 world wide web conference}, pages 913--922.

\bibitem[{Ge et~al.(2022{\natexlab{a}})Ge, Tan, Zhu, Xia, Luo, Liu, Fu, Geng, Li, and Zhang}]{ge2022explainable}
Yingqiang Ge, Juntao Tan, Yan Zhu, Yinglong Xia, Jiebo Luo, Shuchang Liu, Zuohui Fu, Shijie Geng, Zelong Li, and Yongfeng Zhang. 2022{\natexlab{a}}.
\newblock Explainable fairness in recommendation.
\newblock In \emph{Proceedings of the 45th International ACM SIGIR Conference on Research and Development in Information Retrieval}, pages 681--691.

\bibitem[{Ge et~al.(2020)Ge, Zhao, Zhou, Pei, Sun, Ou, and Zhang}]{echochamber}
Yingqiang Ge, Shuya Zhao, Honglu Zhou, Changhua Pei, Fei Sun, Wenwu Ou, and Yongfeng Zhang. 2020.
\newblock Understanding echo chambers in e-commerce recommender systems.
\newblock In \emph{Proceedings of the 43rd international ACM SIGIR conference on research and development in information retrieval}, pages 2261--2270.

\bibitem[{Ge et~al.(2022{\natexlab{b}})Ge, Zhao, Yu, Paul, Hu, Hsieh, and Zhang}]{rl-rec4}
Yingqiang Ge, Xiaoting Zhao, Lucia Yu, Saurabh Paul, Diane Hu, Chu-Cheng Hsieh, and Yongfeng Zhang. 2022{\natexlab{b}}.
\newblock Toward pareto efficient fairness-utility trade-off in recommendation through reinforcement learning.
\newblock In \emph{Proceedings of the fifteenth ACM international conference on web search and data mining}, pages 316--324.

\bibitem[{Geng et~al.(2022)Geng, Liu, Fu, Ge, and Zhang}]{p5}
Shijie Geng, Shuchang Liu, Zuohui Fu, Yingqiang Ge, and Yongfeng Zhang. 2022.
\newblock Recommendation as language processing (rlp): A unified pretrain, personalized prompt \& predict paradigm (p5).
\newblock In \emph{Proceedings of the 16th ACM Conference on Recommender Systems}, pages 299--315.

\bibitem[{Grisse(2023)}]{manupilate}
Karina Grisse. 2023.
\newblock \emph{Recommender Systems, Manipulation and Private Autonomy: How European Civil Law Regulates and Should Regulate Recommender Systems for the Benefit of Private Autonomy}, pages 101--128.
\newblock Springer International Publishing, Cham.

\bibitem[{Gunter et~al.(2024)Gunter, Wang, Wang, Pang, Narayanan, Zhang, Zhang, Chen, Chiu, Qiu et~al.}]{apple2.5b}
Tom Gunter, Zirui Wang, Chong Wang, Ruoming Pang, Andy Narayanan, Aonan Zhang, Bowen Zhang, Chen Chen, Chung-Cheng Chiu, David Qiu, et~al. 2024.
\newblock Apple intelligence foundation language models.
\newblock \emph{arXiv preprint arXiv:2407.21075}.

\bibitem[{Gur et~al.(2023)Gur, Furuta, Huang, Safdari, Matsuo, Eck, and Faust}]{webagent}
Izzeddin Gur, Hiroki Furuta, Austin Huang, Mustafa Safdari, Yutaka Matsuo, Douglas Eck, and Aleksandra Faust. 2023.
\newblock A real-world webagent with planning, long context understanding, and program synthesis.
\newblock \emph{arXiv preprint arXiv:2307.12856}.

\bibitem[{Hidasi et~al.(2015)Hidasi, Karatzoglou, Baltrunas, and Tikk}]{gru4rec}
Bal{\'a}zs Hidasi, Alexandros Karatzoglou, Linas Baltrunas, and Domonkos Tikk. 2015.
\newblock Session-based recommendations with recurrent neural networks.
\newblock \emph{arXiv preprint arXiv:1511.06939}.

\bibitem[{Huang et~al.(2023{\natexlab{a}})Huang, Yu, Ma, Zhong, Feng, Wang, Chen, Peng, Feng, Qin et~al.}]{huang2023survey}
Lei Huang, Weijiang Yu, Weitao Ma, Weihong Zhong, Zhangyin Feng, Haotian Wang, Qianglong Chen, Weihua Peng, Xiaocheng Feng, Bing Qin, et~al. 2023{\natexlab{a}}.
\newblock A survey on hallucination in large language models: Principles, taxonomy, challenges, and open questions.
\newblock \emph{arXiv preprint arXiv:2311.05232}.

\bibitem[{Huang et~al.(2023{\natexlab{b}})Huang, Lian, Lei, Yao, Lian, and Xie}]{huang2023recommender}
Xu~Huang, Jianxun Lian, Yuxuan Lei, Jing Yao, Defu Lian, and Xing Xie. 2023{\natexlab{b}}.
\newblock Recommender ai agent: Integrating large language models for interactive recommendations.
\newblock \emph{arXiv preprint arXiv:2308.16505}.

\bibitem[{J{\"a}rvelin and Kek{\"a}l{\"a}inen(2002)}]{NDCG}
Kalervo J{\"a}rvelin and Jaana Kek{\"a}l{\"a}inen. 2002.
\newblock Cumulated gain-based evaluation of ir techniques.
\newblock \emph{ACM Transactions on Information Systems (TOIS)}, 20(4):422--446.

\bibitem[{Jiang et~al.(2019)Jiang, Chiappa, Lattimore, Gy{\"o}rgy, and Kohli}]{chamberother2}
Ray Jiang, Silvia Chiappa, Tor Lattimore, Andr{\'a}s Gy{\"o}rgy, and Pushmeet Kohli. 2019.
\newblock Degenerate feedback loops in recommender systems.
\newblock In \emph{Proceedings of the 2019 AAAI/ACM Conference on AI, Ethics, and Society}, pages 383--390.

\bibitem[{Kalimeris et~al.(2021)Kalimeris, Bhagat, Kalyanaraman, and Weinsberg}]{chamberotherkdd}
Dimitris Kalimeris, Smriti Bhagat, Shankar Kalyanaraman, and Udi Weinsberg. 2021.
\newblock Preference amplification in recommender systems.
\newblock In \emph{Proceedings of the 27th ACM SIGKDD Conference on Knowledge Discovery \& Data Mining}, pages 805--815.

\bibitem[{Kaminskas and Bridge(2016)}]{polar2}
Marius Kaminskas and Derek Bridge. 2016.
\newblock Diversity, serendipity, novelty, and coverage: a survey and empirical analysis of beyond-accuracy objectives in recommender systems.
\newblock \emph{ACM Transactions on Interactive Intelligent Systems (TiiS)}, 7(1):1--42.

\bibitem[{Kang and McAuley(2018)}]{sasrec}
Wang-Cheng Kang and Julian McAuley. 2018.
\newblock Self-attentive sequential recommendation.
\newblock In \emph{2018 IEEE international conference on data mining (ICDM)}, pages 197--206. IEEE.

\bibitem[{Kunaver and Po{\v{z}}rl(2017)}]{polar3}
Matev{\v{z}} Kunaver and Toma{\v{z}} Po{\v{z}}rl. 2017.
\newblock Diversity in recommender systems--a survey.
\newblock \emph{Knowledge-based systems}, 123:154--162.

\bibitem[{Li et~al.(2019)Li, Liu, Wu, Xu, Zhao, Huang, Kang, Chen, Li, and Lee}]{MIND}
Chao Li, Zhiyuan Liu, Mengmeng Wu, Yuchi Xu, Huan Zhao, Pipei Huang, Guoliang Kang, Qiwei Chen, Wei Li, and Dik~Lun Lee. 2019.
\newblock Multi-interest network with dynamic routing for recommendation at tmall.
\newblock In \emph{Proceedings of the 28th ACM international conference on information and knowledge management}, pages 2615--2623.

\bibitem[{Li et~al.(2023)Li, Chen, Zhao, Zhang, and Xing}]{e4srec}
Xinhang Li, Chong Chen, Xiangyu Zhao, Yong Zhang, and Chunxiao Xing. 2023.
\newblock E4srec: An elegant effective efficient extensible solution of large language models for sequential recommendation.
\newblock \emph{arXiv preprint arXiv:2312.02443}.

\bibitem[{Li et~al.(2021)Li, Chen, Fu, Ge, and Zhang}]{fairnessrec}
Yunqi Li, Hanxiong Chen, Zuohui Fu, Yingqiang Ge, and Yongfeng Zhang. 2021.
\newblock User-oriented fairness in recommendation.
\newblock In \emph{Proceedings of the web conference 2021}, pages 624--632.

\bibitem[{Liu et~al.(2024{\natexlab{a}})Liu, Lin, Hewitt, Paranjape, Bevilacqua, Petroni, and Liang}]{llmbias}
Nelson~F Liu, Kevin Lin, John Hewitt, Ashwin Paranjape, Michele Bevilacqua, Fabio Petroni, and Percy Liang. 2024{\natexlab{a}}.
\newblock Lost in the middle: How language models use long contexts.
\newblock \emph{Transactions of the Association for Computational Linguistics}, 12:157--173.

\bibitem[{Liu et~al.(2024{\natexlab{b}})Liu, Lin, Hewitt, Paranjape, Bevilacqua, Petroni, and Liang}]{liu2024lost}
Nelson~F Liu, Kevin Lin, John Hewitt, Ashwin Paranjape, Michele Bevilacqua, Fabio Petroni, and Percy Liang. 2024{\natexlab{b}}.
\newblock Lost in the middle: How language models use long contexts.
\newblock \emph{Transactions of the Association for Computational Linguistics}, 12:157--173.

\bibitem[{M{\"o}ller et~al.(2020)M{\"o}ller, Trilling, Helberger, and van Es}]{chamberother3}
Judith M{\"o}ller, Damian Trilling, Natali Helberger, and Bram van Es. 2020.
\newblock Do not blame it on the algorithm: an empirical assessment of multiple recommender systems and their impact on content diversity.
\newblock In \emph{Digital media, political polarization and challenges to democracy}, pages 45--63. Routledge.

\bibitem[{Ni et~al.(2019)Ni, Li, and McAuley}]{amazon18dataset}
Jianmo Ni, Jiacheng Li, and Julian McAuley. 2019.
\newblock Justifying recommendations using distantly-labeled reviews and fine-grained aspects.
\newblock In \emph{Proceedings of the 2019 conference on empirical methods in natural language processing and the 9th international joint conference on natural language processing (EMNLP-IJCNLP)}, pages 188--197.

\bibitem[{Park et~al.(2023)Park, O'Brien, Cai, Morris, Liang, and Bernstein}]{park2023generative}
Joon~Sung Park, Joseph O'Brien, Carrie~Jun Cai, Meredith~Ringel Morris, Percy Liang, and Michael~S Bernstein. 2023.
\newblock Generative agents: Interactive simulacra of human behavior.
\newblock In \emph{Proceedings of the 36th annual acm symposium on user interface software and technology}, pages 1--22.

\bibitem[{Patro et~al.(2020)Patro, Biswas, Ganguly, Gummadi, and Chakraborty}]{patro2020fairrec}
Gourab~K Patro, Arpita Biswas, Niloy Ganguly, Krishna~P Gummadi, and Abhijnan Chakraborty. 2020.
\newblock Fairrec: Two-sided fairness for personalized recommendations in two-sided platforms.
\newblock In \emph{Proceedings of the web conference 2020}, pages 1194--1204.

\bibitem[{Qu et~al.(2019)Qu, Yang, Qiu, Croft, Zhang, and Iyyer}]{qu2019bert}
Chen Qu, Liu Yang, Minghui Qiu, W~Bruce Croft, Yongfeng Zhang, and Mohit Iyyer. 2019.
\newblock Bert with history answer embedding for conversational question answering.
\newblock In \emph{Proceedings of the 42nd international ACM SIGIR conference on research and development in information retrieval}, pages 1133--1136.

\bibitem[{Ren and Huang(2024)}]{ren2024easyrec}
Xubin Ren and Chao Huang. 2024.
\newblock Easyrec: Simple yet effective language models for recommendation.
\newblock \emph{arXiv preprint arXiv:2408.08821}.

\bibitem[{Robertson et~al.(2009)Robertson, Zaragoza et~al.}]{bm25}
Stephen Robertson, Hugo Zaragoza, et~al. 2009.
\newblock The probabilistic relevance framework: Bm25 and beyond.
\newblock \emph{Foundations and Trends{\textregistered} in Information Retrieval}, 3(4):333--389.

\bibitem[{Sarwar et~al.(2001)Sarwar, Karypis, Konstan, and Riedl}]{MRR}
Badrul Sarwar, George Karypis, Joseph Konstan, and John Riedl. 2001.
\newblock Item-based collaborative filtering recommendation algorithms.
\newblock In \emph{Proceedings of the 10th international conference on World Wide Web}, pages 285--295.

\bibitem[{Shanahan et~al.(2023)Shanahan, McDonell, and Reynolds}]{shanahan2023role}
Murray Shanahan, Kyle McDonell, and Laria Reynolds. 2023.
\newblock Role play with large language models.
\newblock \emph{Nature}, 623(7987):493--498.

\bibitem[{Sun et~al.(2019)Sun, Liu, Wu, Pei, Lin, Ou, and Jiang}]{bert4rec}
Fei Sun, Jun Liu, Jian Wu, Changhua Pei, Xiao Lin, Wenwu Ou, and Peng Jiang. 2019.
\newblock Bert4rec: Sequential recommendation with bidirectional encoder representations from transformer.
\newblock In \emph{Proceedings of the 28th ACM international conference on information and knowledge management}, pages 1441--1450.

\bibitem[{Sun and Zhang(2018)}]{2018CRS}
Yueming Sun and Yi~Zhang. 2018.
\newblock Conversational recommender system.
\newblock In \emph{The 41st international acm sigir conference on research \& development in information retrieval}, pages 235--244.

\bibitem[{Team et~al.(2024)Team, Mesnard, Hardin, Dadashi, Bhupatiraju, Pathak, Sifre, Rivi{\`e}re, Kale, Love et~al.}]{gemma2b}
Gemma Team, Thomas Mesnard, Cassidy Hardin, Robert Dadashi, Surya Bhupatiraju, Shreya Pathak, Laurent Sifre, Morgane Rivi{\`e}re, Mihir~Sanjay Kale, Juliette Love, et~al. 2024.
\newblock Gemma: Open models based on gemini research and technology.
\newblock \emph{arXiv preprint arXiv:2403.08295}.

\bibitem[{Vaswani(2017)}]{transformer}
A~Vaswani. 2017.
\newblock Attention is all you need.
\newblock \emph{Advances in Neural Information Processing Systems}.

\bibitem[{Wan et~al.(2019)Wan, Misra, Nakashole, and McAuley}]{goodreads}
Mengting Wan, Rishabh Misra, Ndapa Nakashole, and Julian McAuley. 2019.
\newblock Fine-grained spoiler detection from large-scale review corpora.
\newblock \emph{arXiv preprint arXiv:1905.13416}.

\bibitem[{Wang et~al.(2020)Wang, Fan, Xia, Zhao, Niu, and Huang}]{rl-rec3}
Pengfei Wang, Yu~Fan, Long Xia, Wayne~Xin Zhao, ShaoZhang Niu, and Jimmy Huang. 2020.
\newblock Kerl: A knowledge-guided reinforcement learning model for sequential recommendation.
\newblock In \emph{Proceedings of the 43rd International ACM SIGIR conference on research and development in Information Retrieval}, pages 209--218.

\bibitem[{Wang et~al.(2023)Wang, Jiang, Chen, Yang, Zhou, Cho, Fan, Huang, Lu, and Yang}]{wang2023recmind}
Yancheng Wang, Ziyan Jiang, Zheng Chen, Fan Yang, Yingxue Zhou, Eunah Cho, Xing Fan, Xiaojiang Huang, Yanbin Lu, and Yingzhen Yang. 2023.
\newblock Recmind: Large language model powered agent for recommendation.
\newblock \emph{arXiv preprint arXiv:2308.14296}.

\bibitem[{Wang et~al.(2024)Wang, Yu, Zheng, Ma, and Zhang}]{wang2024macrec}
Zhefan Wang, Yuanqing Yu, Wendi Zheng, Weizhi Ma, and Min Zhang. 2024.
\newblock Macrec: A multi-agent collaboration framework for recommendation.
\newblock In \emph{Proceedings of the 47th International ACM SIGIR Conference on Research and Development in Information Retrieval}, pages 2760--2764.

\bibitem[{Xiao et~al.(2023)Xiao, Liu, Zhang, and Muennighoff}]{bge_embedding}
Shitao Xiao, Zheng Liu, Peitian Zhang, and Niklas Muennighoff. 2023.
\newblock \href {https://arxiv.org/abs/2309.07597} {C-pack: Packaged resources to advance general chinese embedding}.
\newblock \emph{Preprint}, arXiv:2309.07597.

\bibitem[{Xie et~al.(2024)Xie, Zhang, Chen, Zhu, Lou, Tian, Xiao, and Su}]{travelagent}
Jian Xie, Kai Zhang, Jiangjie Chen, Tinghui Zhu, Renze Lou, Yuandong Tian, Yanghua Xiao, and Yu~Su. 2024.
\newblock Travelplanner: A benchmark for real-world planning with language agents.
\newblock \emph{arXiv preprint arXiv:2402.01622}.

\bibitem[{Xie et~al.(2021)Xie, Ling, Wang, Wang, Xia, and Lin}]{xie2021deep}
Ruobing Xie, Cheng Ling, Yalong Wang, Rui Wang, Feng Xia, and Leyu Lin. 2021.
\newblock Deep feedback network for recommendation.
\newblock In \emph{Proceedings of the twenty-ninth international conference on international joint conferences on artificial intelligence}, pages 2519--2525.

\bibitem[{Xu et~al.(2022)Xu, Tan, Fu, Ji, Heinecke, and Zhang}]{echochamber2}
Shuyuan Xu, Juntao Tan, Zuohui Fu, Jianchao Ji, Shelby Heinecke, and Yongfeng Zhang. 2022.
\newblock Dynamic causal collaborative filtering.
\newblock In \emph{Proceedings of the 31st ACM International Conference on Information \& Knowledge Management}, pages 2301--2310.

\bibitem[{Xu et~al.(2023)Xu, Li, Ha, Guo, Ma, Liu, Chen, and Zhu}]{nmcdrhead}
Wujiang Xu, Shaoshuai Li, Mingming Ha, Xiaobo Guo, Qiongxu Ma, Xiaolei Liu, Linxun Chen, and Zhenfeng Zhu. 2023.
\newblock Neural node matching for multi-target cross domain recommendation.
\newblock In \emph{2023 IEEE 39th International Conference on Data Engineering (ICDE)}, pages 2154--2166. IEEE.

\bibitem[{Xu et~al.(2024)Xu, Liang, Han, Ning, Lin, Chen, Wei, and Zhang}]{slmrec}
Wujiang Xu, Zujie Liang, Jiaojiao Han, Xuying Ning, Wenfang Lin, Linxun Chen, Feng Wei, and Yongfeng Zhang. 2024.
\newblock Slmrec: Empowering small language models for sequential recommendation.
\newblock \emph{arXiv preprint arXiv:2405.17890}.

\bibitem[{Yao et~al.(2022)Yao, Chen, Yang, and Narasimhan}]{webshop}
Shunyu Yao, Howard Chen, John Yang, and Karthik Narasimhan. 2022.
\newblock Webshop: Towards scalable real-world web interaction with grounded language agents.
\newblock \emph{Advances in Neural Information Processing Systems}, 35:20744--20757.

\bibitem[{Zhai et~al.(2024)Zhai, Liao, Liu, Wang, Li, Cao, Gao, Gong, Gu, He et~al.}]{metaRec}
Jiaqi Zhai, Lucy Liao, Xing Liu, Yueming Wang, Rui Li, Xuan Cao, Leon Gao, Zhaojie Gong, Fangda Gu, Michael He, et~al. 2024.
\newblock Actions speak louder than words: Trillion-parameter sequential transducers for generative recommendations.
\newblock \emph{arXiv preprint arXiv:2402.17152}.

\bibitem[{Zhang et~al.(2024{\natexlab{a}})Zhang, Chen, Sheng, Wang, and Chua}]{generativerecagent}
An~Zhang, Yuxin Chen, Leheng Sheng, Xiang Wang, and Tat-Seng Chua. 2024{\natexlab{a}}.
\newblock On generative agents in recommendation.
\newblock In \emph{Proceedings of the 47th International ACM SIGIR Conference on Research and Development in Information Retrieval}, pages 1807--1817.

\bibitem[{Zhang et~al.(2024{\natexlab{b}})Zhang, Hou, Xie, Sun, McAuley, Zhao, Lin, and Wen}]{agentcf}
Junjie Zhang, Yupeng Hou, Ruobing Xie, Wenqi Sun, Julian McAuley, Wayne~Xin Zhao, Leyu Lin, and Ji-Rong Wen. 2024{\natexlab{b}}.
\newblock Agentcf: Collaborative learning with autonomous language agents for recommender systems.
\newblock In \emph{Proceedings of the ACM on Web Conference 2024}, pages 3679--3689.

\bibitem[{Zhang(2018)}]{zhang2018personalizing}
Saizheng Zhang. 2018.
\newblock Personalizing dialogue agents: I have a dog, do you have pets too.
\newblock \emph{arXiv preprint arXiv:1801.07243}.

\bibitem[{Zhang et~al.(2018)Zhang, Chen, Ai, Yang, and Croft}]{zhang2018CRS}
Yongfeng Zhang, Xu~Chen, Qingyao Ai, Liu Yang, and W~Bruce Croft. 2018.
\newblock Towards conversational search and recommendation: System ask, user respond.
\newblock In \emph{Proceedings of the 27th acm international conference on information and knowledge management}, pages 177--186.

\bibitem[{Zhang et~al.(2014)Zhang, Lai, Zhang, Zhang, Liu, and Ma}]{zhang2014explicit}
Yongfeng Zhang, Guokun Lai, Min Zhang, Yi~Zhang, Yiqun Liu, and Shaoping Ma. 2014.
\newblock Explicit factor models for explainable recommendation based on phrase-level sentiment analysis.
\newblock In \emph{Proceedings of the 37th international ACM SIGIR conference on Research \& development in information retrieval}, pages 83--92.

\bibitem[{Zhao et~al.(2024)Zhao, Wu, Wang, Tang, Wang, and de~Rijke}]{toolrec}
Yuyue Zhao, Jiancan Wu, Xiang Wang, Wei Tang, Dingxian Wang, and Maarten de~Rijke. 2024.
\newblock Let me do it for you: Towards llm empowered recommendation via tool learning.
\newblock In \emph{Proceedings of the 47th International ACM SIGIR Conference on Research and Development in Information Retrieval}, pages 1796--1806.

\bibitem[{Zheng et~al.(2018)Zheng, Zhang, Zheng, Xiang, Yuan, Xie, and Li}]{rl-rec2}
Guanjie Zheng, Fuzheng Zhang, Zihan Zheng, Yang Xiang, Nicholas~Jing Yuan, Xing Xie, and Zhenhui Li. 2018.
\newblock Drn: A deep reinforcement learning framework for news recommendation.
\newblock In \emph{Proceedings of the 2018 world wide web conference}, pages 167--176.

\bibitem[{Zhou et~al.(2019)Zhou, Mou, Fan, Pi, Bian, Zhou, Zhu, and Gai}]{DIEN}
Guorui Zhou, Na~Mou, Ying Fan, Qi~Pi, Weijie Bian, Chang Zhou, Xiaoqiang Zhu, and Kun Gai. 2019.
\newblock Deep interest evolution network for click-through rate prediction.
\newblock In \emph{Proceedings of the AAAI conference on artificial intelligence}, volume~33, pages 5941--5948.

\bibitem[{Zhou et~al.(2018)Zhou, Zhu, Song, Fan, Zhu, Ma, Yan, Jin, Li, and Gai}]{DIN}
Guorui Zhou, Xiaoqiang Zhu, Chenru Song, Ying Fan, Han Zhu, Xiao Ma, Yanghui Yan, Junqi Jin, Han Li, and Kun Gai. 2018.
\newblock Deep interest network for click-through rate prediction.
\newblock In \emph{Proceedings of the 24th ACM SIGKDD international conference on knowledge discovery \& data mining}, pages 1059--1068.

\end{thebibliography}

% \newpage

\clearpage
% \section*{APPENDIX}
\onecolumn
\tableofcontents

\section*{APPENDIX}
% \clearpage
\appendix

\section{Task Definitions and Comparisions}~\label{sec:definitions}
% \noindent \textbf{Definition of Related Tasks.}

\noindent \textbf{Sequential Recommendation.}  Consider a set of users \( U \) and a set of items \( I \). Each user's historical interactions are represented by a sequence \( S_u = [s_1, \dots, s_i, \dots, s_T] \), where \( s_i \in I \) and \( T \) is the length of the sequence. The goal of sequential recommendation is to predict the next item \( s_{T+1} \) that the user \( u \) is likely to interact with, based on their past interactions \( S_u \)~\citep{gru4rec,sasrec,bert4rec,p5}.
Formally, this involves estimating the probability distribution over the items for the next interaction:
\begin{equation}
  \hat{i} = \arg\max_{i \in I} P(s_{T+1} = i \mid S_u; \psi).
\end{equation}
where $\psi$ is the model's parameters.
Recent work on recommendation agents~\citep{agentcf,wang2023recmind,generativerecagent} has leveraged large language models (LLMs) to simulate user behavior by prompting them with plain text descriptions of user history and learn from the external knowledge via tool usage. Despite the shift to a language-based framework, it shares the same optimization objective as the traditional sequential recommendation.

\noindent \textbf{Conversational Recommendation.} Traditional conversational recommendation system~\citep{2018CRS,zhang2018CRS} analyzes the user's intention via the multiple turn dialogue and consider historical information to achieve personalized recommendation. Mathematically, the recommendation model part\footnote{The conversational model part is omitted for concise.} can be summarized as: 
% 1. LLM background
\begin{equation}
\hat{i} = \arg\max_{i \in I} P(s_{T+1} = i \mid S_u, H_u; \psi).
\end{equation}
where $H_u=[h_1,...,h_R]$ represents multiple historical dialogues of a user, $R$ represents the number of dialogues and $\psi$ is the model's parameters.
% 1. plan - fixed of CRS, instruction-invariant | instruction-specific. citation
% 2. agent instruction | flexible instruction, CRS:
% 3. tool using - agent

\noindent \textbf{Our Task.} Unlike sequential and conversational recommendation, our task focuses on learning from user's instructions to build an agentic shield between user and recommender system and meanwhile provide personalized recommendations for users. 
% Our models are designed to acquire external world knowledge based on user-guided instructions. 
Mathematically, this can be summarized as follows:
\begin{equation}
\hat{i} = \arg\max_{i \in I} P(s_{T+1} = i \mid S_u, \Omega_u, E ; \psi_u).
\end{equation}
where $\Omega_u$ represents the user's instructions, and $\psi_u$ denotes the user-specific model parameters. $E$ represents the external environment, which can supply real-time information to the agent.

In Table~\ref{tab:diff}, we highlight the key differences between previous recommendation models and our proposed model. Unlike existing recommendation models, our approach conducts an in-depth analysis of users' instructions and learns from individual feedback. Additionally, leveraging the power of LLMs, our model supports a highly flexible range of instructions and dialogues.
% As shown in Table \ref{tab:diff}, 

% \begin{table}[ht]
% \centering
% \caption{Difference between previous recommendation models and this task}
% \resizebox{0.9\textwidth}{!}{
% \begin{tabular}{c|c|c|c|c}
% \toprule
% Task  & \makecell[c]{Sequential \\ Recommendation} & \makecell[c]{Conversational \\ Recommendation} & \makecell[c]{Recommendation \\ Agent} & Ours \\
% \midrule
% Instruction Awareness  & \xmark  & \cmark  & \xmark  &  \cmark \\
% Instruction Type  & N/A  & Fixed  & N/A &  Flexible \\
% Dialogue Interaction  & N/A  & Multiple Turns  & N/A  & 0, 1, or Multiple Turns \\
% Dynamic Interest   & \xmark  & \cmark  & \xmark  &  \cmark \\
% Personalized Model  & \xmark  & \xmark  & \xmark  &  \cmark \\
% External Knowledge  & \xmark  & \xmark  & \cmark  &  \cmark \\
% \bottomrule
% \end{tabular}
% }
% \label{tab:diff}
% \end{table}

\begin{table*}[tb!]
\centering
\caption{Difference between previous recommendation models and our model.}
\vspace{-1.0em}
\resizebox{0.98\textwidth}{!}{
\begin{tabular}{c|c|c|c|c|c|c}
\midrule
\textbf{Model} & \textbf{\makecell[c]{Instruction \\ Awareness}} & \textbf{\makecell[c]{Instruction \\ Type}} & \textbf{\makecell[c]{Dialogue \\ Interaction}} & \textbf{\makecell[c]{Dynamic \\ Interest}} & \textbf{\makecell[c]{Learning \\ from Feedback}} & \textbf{\makecell[c]{External \\ Knowledge}} \\
\midrule
SR & \xmark & N/A & N/A & \xmark & \xmark & \xmark \\
CRS & \cmark & Fixed & Multiple Turns & \cmark & \xmark & \xmark \\
RecAgent & \xmark & N/A & N/A & \xmark & \xmark & \cmark \\
\midrule
Ours & \cmark & Flexible & 0, 1, or Multiple Turns & \cmark & \cmark & \cmark \\
\midrule
\end{tabular}
}
\label{tab:diff}
\vspace{-5pt}
\end{table*}

% \appendix
% \centerline{\maketitle{\textbf{SUMMARY OF THE APPENDIX}}}
% \section{Appendix}

% \section{Details of Dataset Construction}
\section{Detailed Related Work}~\label{sec:appendix:related work}

\subsection{Recommender System}
% first introduce CRS - on hold

Sequential recommendation models~\citep{gru4rec,bert4rec,sasrec} primarily focus on developing temporal encoders to capture both short- and long-term user interests. For instance, SASRec~\citep{sasrec} leverages an attention mechanism to capture long-term semantics, while BERT4Rec~\citep{bert4rec} uses a bidirectional encoder with a masked item training objective. In the context of embracing large language models, generative recommenders~\citep{p5,metaRec} treat item indices as tokens and predict them in a generative manner. Meanwhile, LLMs~\citep{e4srec,slmrec} are utilized to play as a sequential embedding extractor to improve the recommendation performance. In our framework design, all recommendation models can be considered as components of the tools.

Before large language model become popular, conversational recommendation system (CRS)~\citep{2018CRS,zhang2018CRS,qu2019bert} aims at designing better dialogue understanding models or incorporating reinforcement learning for multiple dialogues answering. Due to the capacity of the conventional language model, it lose the flexibility of the dialogue including the dialogue format and number of turns. To resolve this problem, some researchers~\citep{LLM+CRSgoogle,LLM+CRS2} leverage the power of LLM to better understand the intention of user. 

% echo chamber based on the popularity and filtering of our work | introduce fairness yunqi to introduce to protect  | from protecting of irrevalant item / long-tailed items / 

The echo chamber effect occurs when individuals are exposed only to information and opinions that reinforce their existing beliefs within their social networks~\citep{echoeffect1,echoeffect2,echoeffect3}, leading to a lack of diverse perspectives and increased polarization~\citep{polar1,polar2,polar3}. In the context of recommender systems, researchers have begun to study echo chambers and feedback loops~\citep{echochamber,echochamber2,chamberother1,chamberother2,chamberother3,chamberotherkdd}. Kalimeris et al.~\citep{chamberotherkdd} propose a matrix factorization-based recommender system with a theoretical framework for modeling dynamic user interests, while $\partial$CCF~\citep{echoeffect2} employs counterfactual reasoning to mitigate echo chambers.

\subsection{Personal Language-based Agent}
In the early stages, some researchers~\citep{zhang2018personalizing, park2023generative, shanahan2023role} in the NLP field developed dialogue agents with personas to enhance dialogue quality. Language models~\citep{park2023generative} are prompted with role descriptions to simulate realistic interactions by storing experiences, synthesizing memories, and dynamically planning actions, resulting in believable individual and social behaviors within interactive environments. 
% webshop mind2web, travel agent or other agents
% recagent
WebShop~\citep{webshop} attempts to understand product attributes from human-provided text instructions using reinforcement learning and imitation learning. Similar to traditional conversational recommender systems (CRS)~\citep{zhang2018CRS}, it is impractical for users to describe each product attribute every time. With the advancement of large language models (such as GPTs~\citep{gpt4}), many researchers~\citep{webagent,mind2web,travelagent} have begun designing domain-specific agents that integrate various tool learning and memory mechanisms.

More recently, recommendation agents (RecAgent)~\citep{toolrec,wang2023recmind,generativerecagent,agentcf,wang2024macrec,huang2023recommender} have been developed to simulate user behaviors and predict user-item interactions. A common design feature among these agents is the use of historical interaction information as user memory~\citep{toolrec,wang2023recmind,huang2023recommender}, with LLMs utilized to generate the ranking results. Unlike platform-side RecAgents, $\model$ and $\modelplus$ are the first to operate on the user side, generating re-ranking results based on user instructions and individual memory, unaffected by the influence of advantaged users.

\section{Experiment}
\label{appendix::experiment results}
\subsection{Source Dataset}
\noindent\textbf{Amazon Book/Movietv}~\footnote{\url{https://cseweb.ucsd.edu/~jmcauley/datasets/amazon_v2/}}~\citep{amazon18dataset} The Amazon product dataset is a comprehensive repository of consumer reviews and associated metadata, encompassing 142.8 million reviews collected over an 18-year span from May 1996 to July 2014. For our experiments, we leverage two distinct subsets: "Books" and "Movies and TV." Each dataset includes anonymized user and item identifiers, along with user-provided ratings on a 1-5 scale and corresponding textual reviews. Furthermore, rich product metadata is incorporated, such as detailed descriptions, categorical classifications, pricing information, and brand data. This multifaceted dataset provides a fertile ground for both collaborative filtering and content-based recommendation approaches, where the interplay between user behavior, product attributes, and textual feedback can be modeled to advance the state of recommendation systems.

\noindent\textbf{Goodreads.}~\footnote{\url{https://mengtingwan.github.io/data/goodreads}}~\citep{goodreads} The Goodreads dataset is derived from one of the largest online platforms dedicated to book reviews, offering user-generated ratings, reviews, and a variety of associated metadata. Each user in the dataset is represented by an anonymized identifier, with interactions including rating and reviewing a broad selection of books. The books are identified through International Standard Book Numbers (ISBNs) and accompanied by an extensive set of metadata, including title, author, publication year, and genre classifications. This data is especially valuable for the development of content-aware recommendation models, where leveraging the contextual features of both user interactions and book attributes can enhance predictive accuracy. The textual reviews, in particular, provide a rich source of natural language data, capturing nuanced user feedback that can be further utilized in sentiment analysis, opinion mining, and advanced NLP tasks. Ratings, similarly to the Amazon dataset, are presented on a 1-5 scale, providing a consistent metric for comparative analysis across different datasets.

\noindent\textbf{Yelp.}~\footnote{\url{https://www.kaggle.com/datasets/yelp-dataset/yelp-dataset/versions}} The Yelp dataset contains over 67,000 reviews focused on businesses, particularly restaurants, from three major English-speaking cities, sourced from the popular Yelp platform. The dataset includes detailed metadata on both businesses and user interactions. Each business is uniquely identified and linked to comprehensive metadata, including its name, geographic location, category (e.g., restaurant, bar, or retail establishment), and additional attributes such as parking availability and reservation policies. This data is invaluable for context-aware recommendation systems, where business features and user feedback intersect to inform personalized recommendations. Anonymized user IDs track user interactions, with additional features such as the number of reviews written, average rating, and social features (e.g., "friends," "useful votes"). Yelp’s textual reviews provide a rich dataset for natural language processing, where the diverse nature of user opinions, combined with structured metadata, offers a robust framework for evaluating and improving context-aware recommendation models.

\subsection{Compared Methods}
\label{appendix:baselines}
\subsubsection{Sequential recommendation
methods}
For the sequential recommendation baselines, only item ID information was considered in the model. To optimize performance, we experimented with various hyperparameters. The embedding dimension was tested across \{32, 64, 128\}, while the hidden representation in the prediction head ranged from \{8, 16, 32\}. Additionally, the learning rate was evaluated with values of \{1e$^{-3}$, 4e$^{-3}$, 1e$^{-4}$, 4e$^{-4}$\}. The best results are reported based on the highest MRR metric on the validation set.

\textbf{GRU4Rec} \citep{gru4rec} addresses the challenge of modeling sparse sequential data while adapting RNN models to recommender systems. The authors propose a new ranking loss function specifically designed for training these models. The PyTorch implementation of GRU4Rec is available at the URL\footnote{\url{https://github.com/hungpthanh/GRU4REC-pytorch}}.

\textbf{BERT4Rec} \citep{bert4rec} introduces a bidirectional self-attention network to model user behavior sequences. To prevent information leakage and optimize training, it employs a Cloze objective to predict randomly masked items by considering both their left and right context. The PyTorch implementation of BERT4Rec can be found at the URL\footnote{\url{https://github.com/jaywonchung/BERT4Rec-VAE-Pytorch}}.

\textbf{SASRec} \citep{sasrec} is a self-attention-based sequential model designed to balance model parsimony and complexity in recommendation systems. Using an attention mechanism, SASRec identifies relevant items in a user's action history and predicts the next item with relatively few actions, while also capturing long-term semantics, similar to RNNs. This allows SASRec to perform well on both sparse and denser datasets. The PyTorch implementation of SASRec is available at the URL\footnote{\url{https://github.com/pmixer/SASRec.pytorch}}.

\subsubsection{Instruction-aware methods}

We treat the concatenated text of the instruction as the query, while each candidate item is represented by its various metadata (e.g., title, description), transformed into textual format. These textual representations of candidate items are treated as individual 'documents,' forming the document corpus that instruction-aware methods rank based on relevance to the query. By leveraging the semantic richness of both the query and item metadata, this approach enables a context-aware ranking system, prioritizing items according to their alignment with the user's intent and preferences as conveyed through the instruction.

\noindent\textbf{BM25.}~\citep{bm25} BM25, a probabilistic ranking function, is a foundational method in information retrieval, widely used to rank documents based on their relevance to a given query. The core concept of BM25 is to measure the similarity between a query and a document by considering both the frequency of query terms within the document and the distribution of those terms across the entire document corpus. BM25 balances two key factors: term frequency, which reflects how often a query term appears in a document (assuming that higher frequency indicates greater relevance), and inverse document frequency, which assigns more weight to rarer terms in the dataset, as they carry greater informational value. The PyTorch implementation of BM25 is available at the URL\footnote{\url{https://github.com/dorianbrown/rank_bm25}}.

\noindent\textbf{BGE-Rerank.}~\citep{bge_embedding} The BGE-Rerank model utilizes a cross-encoder architecture, where both the query and document are processed together as a single input to directly generate a relevance score. Unlike bi-encoder models, which create independent embeddings for the query and document before computing their similarity, the cross-encoder applies full attention over the entire input pair, capturing more fine-grained interactions. This approach leads to higher accuracy in estimating relevance. In our implementation, we use the BGE-Rerank model to reorder candidate documents based on the relevance score for each query-document pair. The PyTorch implementation of BGE-Rerank is available at the URL\footnote{\url{https://github.com/FlagOpen/FlagEmbedding/tree/master/FlagEmbedding/reranker}}.

\noindent\textbf{EasyRec.} EasyRec~\citep{ren2024easyrec} is a lightweight, highly efficient recommendation system based on large language models, shown through extensive evaluations to outperform many LLM-based methods in terms of accuracy. Central to its success is the use of contrastive learning, which effectively aligns semantic representations from textual data with collaborative filtering signals. This approach enables EasyRec to generalize robustly and adapt to new, unseen recommendation data. The model employs a bi-encoder architecture, where text embeddings for queries and documents are pre-computed independently. These embeddings are then used to calculate similarity scores, allowing for the reordering of candidate items based on relevance. The PyTorch implementation of EasyRec is available at the URL\footnote{\url{https://github.com/HKUDS/EasyRec}}.

\subsubsection{Recommendation Agents}

\noindent\textbf{ToolRec.}~\citep{toolrec} uses large language models (LLMs) to enhance recommendation systems by leveraging external tools. The methodology involves treating LLMs as surrogate users, who simulate user decision-making based on preferences and utilize attribute-oriented tools (such as rank and retrieval tools) to explore and refine item recommendations. This iterative process allows for a more fine-grained recommendation that aligns with users' preferences.

\noindent\textbf{AgentCF.}~\citep{agentcf} AgentCF is an innovative approach that constructs both user and item agents, powered by LLMs, to simulate user-item interactions in recommender systems. These agents are equipped with memory modules designed to capture their intrinsic preferences and behavioral data. At its core, AgentCF facilitates autonomous interactions between user and item agents, enabling them to make decisions based on simulated preferences. A key feature of this framework is the collaborative reflection mechanism, through which agents continuously update their memory, thereby improving their capacity to model real-world user-item relationships over time. 

To ensure a fair comparison and optimize computational efficiency, the number of memory-building rounds in AgentCF is set to 1, matching that of our $\modelplus$. In AgentCF's experiments, the dataset size is 100, which represents only around 0.1\% of the size of our dataset. Moreover, to ensure the generated reranking list without hallucination, we also equipped ToolRec and AgentCF with our self-reflection mechanism.
 
% In our study, based on the original framework, we reproduce the system using only user agents. The user personas are learned through a reflection mechanism applied to behavior sequences of length 5.

\subsection{Performance Comparison}

\subsubsection{Echo Chamber Effect} \label{appendix:echo chamber}
We also report the experimental results evaluating the echo chamber effect in Table~\ref{appendix:tab:exper:echo book}, Table~\ref{appendix:tab:exper:echo_movietv_goodreads} and Table~\ref{appendix:tab:exper:echo_yelp}. Ads items are randomly inserted into the candidate ranking list from other domains to simulate advertising scenarios that users may have encountered. To mitigate position bias in LLMs~\citep{llmbias}, Ads items are added randomly within the candidate list positions. $\modelplus$ accurately identifies users' instructions and extracts knowledge about their underlying needs, thereby effectively removing undesired Ads. Benefitting from not being trained in a purely data-driven manner and constructing user profiles based on their feedback, our $\modelplus$ also recommends more diverse items to users (both active and less-active items), instead of focusing solely on popular items, and meanwhile improves the overall recommendation performance. Drawing from these experimental results, we conclude that our $\modelplus$ can mitigate the echo chamber effect and act as a protective shield for users.

\subsubsection{Protect Less-Active Users} \label{appendix:active}
We define the top 20\% of users as active, with the remaining 80\% classified as less-active~\citep{fairnessrec,nmcdrhead}. Since our data is sampled and filtered using a 10-core process, most users exhibit rich behavioral patterns. Consequently, active users tend to show poorer performance compared to less-active users, largely due to the decline in LLM performance with longer texts~\citep{liu2024lost}.
As illustrated in Table~\ref{appendix:tab:exper_active_movie}, Table~\ref{appendix:tab:exper-active-goodreads} and Table~\ref{appendix:tab:exper-active-yelp}, our $\modelplus$ enhances the performance for both active and less-active users. For less-active users, we construct individual profiles based on their feedback, ensuring that these profiles are not influenced by other users. The experimental results demonstrate that our dynamic memory mechanism offers personalized services tailored to each user individually.

\begin{table}[tb!]
		\centering
		\caption{Evaluation effects (\%) of the echo chamber ($\uparrow$) on the $\dataset$-Amazon Books. We highlight the methods with the \textbf{\textcolor{teal}{first}}, \textbf{\textcolor{purple}{second}} and \textbf{third} best performances.}
  \vspace{-5pt}
		\label{appendix:tab:exper:echo book}
		\resizebox{0.98\textwidth}{!}{
		\begin{tabular}{c|cccc|cccc}
		\midrule
		\multirow{2}{*}{\textbf{Model}}  & \multicolumn{4}{c|}{\texttt{Amazon Book}} & \multicolumn{4}{c}{\texttt{Amazon Book}} \\
         & \textbf{FR@1} & \textbf{FR@3} & \textbf{FR@5} & \textbf{FR@10}   & \textbf{P-HR@1} & \textbf{P-HR@3} & \textbf{P-NDCG@3} & \textbf{P-MRR} \ \\
		\midrule
            
            EasyRec  & 68.41	& 64.32	 &60.30	 &0.03	&\textbf{37.60}	&\textbf{59.28}	&\textbf{50.00}	&\textbf{56.09} \\

            ToolRec & \textbf{70.13} &	\textbf{66.61}	&\color{purple}\textbf{62.41}	&0.00	&12.63	&36.74	&26.24	&35.80 \\

            AgentCF  & 58.02	&50.04	&41.32	&\textbf{0.06}	&17.00	&41.10	&30.68	&39.42 \\
            
            \midrule
            
            {\model}  &\color{purple}\textbf{71.98}	&\color{purple}\textbf{67.82}	&\textbf{60.74}	&\color{purple}\textbf{0.08}	&\color{purple}\textbf{38.85}	&\color{purple}\textbf{59.51}	&\color{purple}\textbf{50.70}	&\color{purple}\textbf{57.32} \\
            
            % & \color{purple}\textbf{31.89} 

            {\modelplus}  &\color{teal}\textbf{77.15}	&\color{teal}\textbf{70.15}	&\color{teal}\textbf{64.05}	&\color{teal}\textbf{0.09}	&\color{teal}\textbf{42.62}	&\color{teal}\textbf{64.70}	&\color{teal}\textbf{55.25}	&\color{teal}\textbf{60.87} \\
            % & \color{teal}\textbf{35.11} & \color{teal}\textbf{53.51} & \color{teal}\textbf{45.64} & \color{teal}\textbf{50.28} &  \color{teal}\textbf{46.43} & \color{teal}\textbf{65.77} & \color{teal}\textbf{57.67} & \color{teal}\textbf{60.43} \\

		\midrule
	\end{tabular}				
		}
    \vspace{-20pt}
\end{table}

\begin{table}[tb!]
		\centering
		\caption{Evaluation effects (\%) of the echo chamber ($\uparrow$) on the $\dataset$-Amazon Movietv and $\dataset$-GoodReads. We highlight the methods with the \textbf{\textcolor{teal}{first}}, \textbf{\textcolor{purple}{second}} and \textbf{third} best performances.}
  \vspace{-5pt}
		\label{appendix:tab:exper:echo_movietv_goodreads}
		\resizebox{0.98\textwidth}{!}{
		\begin{tabular}{c|cccc|cccc}
		\midrule
		\multirow{2}{*}{\textbf{Model}}  & \multicolumn{4}{c|}{\texttt{Amazon Movietv}} & \multicolumn{4}{c}{\texttt{GoodReads}} \\
         & \textbf{P-HR@1} & \textbf{P-HR@3} & \textbf{P-NDCG@3} & \textbf{P-MRR}  & \textbf{P-HR@1} & \textbf{P-HR@3} & \textbf{P-NDCG@3} & \textbf{P-MRR} \ \\
		\midrule
            
            EasyRec & \textbf{37.31}	&\color{purple}\textbf{65.45}	&\color{purple}\textbf{53.54}	&\color{purple}\textbf{56.69} 
            &14.22	&35.98	&26.56	&33.84
            \\

            ToolRec &14.73	&38.12	&27.96	&35.57
            &19.21	&43.22	&32.92	&38.88
            \\

            AgentCF  &27.61	&53.33	&42.37	&47.37 
            &\textbf{21.82}	&\textbf{46.62}	&\textbf{35.99}	&\textbf{41.47}\\
            
            \midrule
            
            {\model}  &\color{purple}\textbf{40.50}	&\textbf{60.71}	&\textbf{52.11}	&\textbf{56.61} 
            &\color{purple}\textbf{23.75}	&\color{purple}\textbf{47.50}	&\color{purple}\textbf{37.34}	&\color{purple}\textbf{42.68} \\

            {\modelplus}  &\color{teal}\textbf{49.51}	&\color{teal}\textbf{70.47}	&\color{teal}\textbf{61.67}	&\color{teal}\textbf{64.69} 
            &\color{teal}\textbf{31.22}	&\color{teal}\textbf{57.33}	&\color{teal}\textbf{46.23}	&\color{teal}\textbf{49.71} \\

		\midrule
	\end{tabular}				
		}
    \vspace{-5pt}
\end{table}

\begin{table}[tb!]
		\centering
		\caption{Evaluation effects (\%) of the echo chamber ($\uparrow$) on the $\dataset$-Yelp. We highlight the methods with the \textbf{\textcolor{teal}{first}}, \textbf{\textcolor{purple}{second}} and \textbf{third} best performances. }
  \vspace{-5pt}
		\label{appendix:tab:exper:echo_yelp}
		\resizebox{0.98\textwidth}{!}{
		\begin{tabular}{c|cccc|cccc}
		\midrule
		\multirow{2}{*}{\textbf{Model}}  & \multicolumn{4}{c|}{\texttt{Yelp}} & \multicolumn{4}{c}{\texttt{Yelp}} \\
         & \textbf{FR@1} & \textbf{FR@3} & \textbf{FR@5} & \textbf{FR@10}   & \textbf{P-HR@1} & \textbf{P-HR@3} & \textbf{P-NDCG@3} & \textbf{P-MRR} \ \\
		\midrule
            
            EasyRec  & \textbf{76.45}	&\textbf{66.50}	&\color{purple}\textbf{57.16}	&\textbf{0.05}	&\textbf{37.18}	&\textbf{61.05}	&\textbf{52.51}	&\textbf{56.85} \\

            ToolRec & 72.64	&63.64	&53.29	&0.00 	&12.40	&32.50	&23.88	&32.73 \\

            AgentCF  & 71.30	&64.15	&52.01	&0.02 	&14.73	&38.46	&28.33	&36.44 \\
            
            \midrule
            
            {\model}  & \color{purple}\textbf{78.24}	&\color{purple}\textbf{69.71}	&\textbf{56.17}	&\color{purple}\textbf{0.12}	&\color{purple}\textbf{41.74}	&\color{purple}\textbf{62.74}	&\color{purple}\textbf{53.82}	&\color{purple}\textbf{58.76} \\

            {\modelplus}  & \color{teal}\textbf{87.69}	&\color{teal}\textbf{86.20}	&\color{teal}\textbf{84.00}	&\color{teal}\textbf{0.16}	&\color{teal}\textbf{43.67}	&\color{teal}\textbf{64.48}	&\color{teal}\textbf{55.62}	&\color{teal}\textbf{60.20} \\

		\midrule
	\end{tabular}				
		}
    \vspace{-5pt}
\end{table}

\begin{table}[tb!]
		\centering
		\caption{The performance (\%) of active  and less-active users on  $\dataset$ - Amazon Movietv. We highlight the methods with the \textbf{\textcolor{teal}{first}}, \textbf{\textcolor{purple}{second}} and \textbf{third} best performances. }
  \vspace{-5pt}
		\label{appendix:tab:exper_active_movie}
		\resizebox{0.95\textwidth}{!}{
		\begin{tabular}{c|cccc|cccc}
		\midrule
		\multirow{2}{*}{\textbf{Model}}  & \multicolumn{4}{c|}{\texttt{Less-Active Users}} & \multicolumn{4}{c}{\texttt{Active Users}} \\
         & \textbf{HR@1} & \textbf{HR@3} & \textbf{NDCG@3} & \textbf{MRR} & \textbf{HR@1} & \textbf{HR@3} & \textbf{NDCG@3} & \textbf{MRR} \\
		\midrule

            EasyRec  &\textbf{35.17}	&\color{purple}\textbf{61.56}	&\color{purple}\textbf{50.39}	&\textbf{53.21}	&\color{purple}\textbf{35.47}	&\color{purple}\textbf{63.15}	&\color{purple}\textbf{51.26}	&\color{purple}\textbf{53.64} \\
            
            ToolRec &14.43	&36.56	&26.96	&33.81	&12.98	&32.18	&23.94	&31.79 \\

            AgentCF &27.38	&50.98	&40.91	&45.36	&21.84	&45.58	&35.57	&40.76 \\

            \midrule

            {\model} &\color{purple}\textbf{39.36}	&\textbf{57.85}	&\textbf{49.98}	&\color{purple}\textbf{53.96}	&\textbf{34.95}	&\textbf{55.19}	&\textbf{46.88}	&\textbf{51.02} \\

            {\modelplus}  &\color{teal}\textbf{47.32}	&\color{teal}\textbf{66.64}	&\color{teal}\textbf{58.57}	&\color{teal}\textbf{61.22}	&\color{teal}\textbf{44.71}	&\color{teal}\textbf{64.99}	&\color{teal}\textbf{56.60}	&\color{teal}\textbf{59.30}  \\

            % EasyRec  & \textbf{30.70} & \textbf{48.87} & \textbf{41.09} & \textbf{46.14} & \textbf{34.96} & \textbf{61.30} & \textbf{50.15} & \textbf{52.98} \\

            % ToolRec & 10.56 & 30.60 & 21.88 & 29.77 & 13.84 & 35.67 & 26.20 & 33.21 \\

            % AgentCF  & 14.24 & 34.16 & 25.55 & 32.77 & 25.90 & 49.82 & 39.64 & 44.23 \\
            
            % \midrule
            
            % {\model}  & \color{purple}\textbf{31.89} & \color{purple}\textbf{48.99} & \color{purple}\textbf{41.69} & \color{purple}\textbf{47.23} & \color{purple}\textbf{38.19} & \color{purple}\textbf{56.87} & \color{purple}\textbf{48.93} & \color{purple}\textbf{53.04} \\

            % {\modelplus}  & \color{teal}\textbf{35.11} & \color{teal}\textbf{53.51} & \color{teal}\textbf{45.64} & \color{teal}\textbf{50.28} &  \color{teal}\textbf{46.43} & \color{teal}\textbf{65.77} & \color{teal}\textbf{57.67} & \color{teal}\textbf{60.43} \\

		\midrule
	\end{tabular}				
		}
    \vspace{-5pt}
\end{table}

\begin{table}[tb!]
		\centering
		\caption{The performance (\%) of active  and less-active users on  $\dataset$ - GoodReads. We highlight the methods with the \textbf{\textcolor{teal}{first}}, \textbf{\textcolor{purple}{second}} and \textbf{third} best performances.}
  \vspace{-5pt}
		\label{appendix:tab:exper-active-goodreads}
		\resizebox{0.95\textwidth}{!}{
		\begin{tabular}{c|cccc|cccc}
		\midrule
		\multirow{2}{*}{\textbf{Model}}  & \multicolumn{4}{c|}{\texttt{Less-Active Users}} & \multicolumn{4}{c}{\texttt{Active Users}} \\
         & \textbf{HR@1} & \textbf{HR@3} & \textbf{NDCG@3} & \textbf{MRR} & \textbf{HR@1} & \textbf{HR@3} & \textbf{NDCG@3} & \textbf{MRR} \\
		\midrule
            
            EasyRec  & 14.44	&35.77	&26.55	&33.67	&14.13	&36.86	&27.09	&33.86 \\

            ToolRec & 19.85	 &43.34	& 33.29 &	39.11	&17.89	&42.02	&31.63	&37.35 \\

            AgentCF  & \textbf{22.91}	&\textbf{46.67}	&\textbf{36.50}	&\textbf{41.89}	&\textbf{19.82}	&\textbf{46.70}	&\textbf{35.22}	&\textbf{40.10} \\
            
            \midrule
            
            {\model}  & \color{purple}\textbf{24.57}	&\color{purple}\textbf{48.12}	&\color{purple}\textbf{38.00}	&\color{purple}\textbf{43.04}	&\color{purple}\textbf{22.62}	&\color{purple}\textbf{46.96}	&\color{purple}\textbf{36.64}	&\color{purple}\textbf{41.70} \\

            {\modelplus}  & \color{teal}\textbf{32.67}	&\color{teal}\textbf{58.08}	&\color{teal}\textbf{47.28}	&\color{teal}\textbf{50.46}	&\color{teal}\textbf{29.76}	&\color{teal}\textbf{55.39}	&\color{teal}\textbf{44.56}	&\color{teal}\textbf{48.19} \\

		\midrule
	\end{tabular}				
		}
    \vspace{-5pt}
\end{table}

\begin{table}[tb!]
		\centering
		\caption{The performance (\%) of active  and less-active users on  $\dataset$ - Yelp. We highlight the methods with the \textbf{\textcolor{teal}{first}}, \textbf{\textcolor{purple}{second}} and \textbf{third} best performances.}
  \vspace{-5pt}
		\label{appendix:tab:exper-active-yelp}
		\resizebox{0.95\textwidth}{!}{
		\begin{tabular}{c|cccc|cccc}
		\midrule
		\multirow{2}{*}{\textbf{Model}}  & \multicolumn{4}{c|}{\texttt{Less-Active Users}} & \multicolumn{4}{c}{\texttt{Active Users}} \\
         & \textbf{HR@1} & \textbf{HR@3} & \textbf{NDCG@3} & \textbf{MRR} & \textbf{HR@1} & \textbf{HR@3} & \textbf{NDCG@3} & \textbf{MRR} \\
		\midrule
            
            EasyRec  & \textbf{32.83}	&\color{purple}\textbf{56.50}	&\textbf{46.29}	&\textbf{50.13}	&\textbf{30.17}	&\textbf{50.87}	&\textbf{42.03}	&\textbf{47.16} \\

            ToolRec & 11.79	&31.21	&22.88	&30.14	&14.21	&32.42	&24.66	&32.11 \\

            AgentCF  & 13.11	&34.72	&25.50	&32.46	&13.22	&36.41	&26.45	&32.89 \\
            
            \midrule
            
            {\model}  & \color{purple}\textbf{37.80}	&\textbf{56.17}	&\color{purple}\textbf{48.37}	&\color{purple}\textbf{52.70}	&\color{purple}\textbf{39.40}	&\color{teal}\textbf{59.10}	&\color{purple}\textbf{50.62}	&\color{purple}\textbf{53.90} \\

            {\modelplus}  & \color{teal}\textbf{39.02}	&\color{teal}\textbf{58.49}	&\color{teal}\textbf{50.23}	&\color{teal}\textbf{53.88}	&\color{teal}\textbf{43.25}	&\color{purple}\textbf{57.75}	&\color{teal}\textbf{51.48}	&\color{teal}\textbf{56.05} \\

		\midrule
	\end{tabular}				
		}
    \vspace{-5pt}
\end{table}
\section{Prompt Templates and Examples}
\label{appendix::prompt template}
\vspace{-5pt}
All output messages are decoded in a JSON-structured format through the OpenAI service~\footnote{\url{https://platform.openai.com/docs/guides/structured-outputs/introduction}}.
\vspace{-5pt}
\subsection{Prompt Templates and Examples Response in $\model$}
\vspace{-5pt}
\subsubsection{Parser}
\vspace{-5pt}
\textbf{With the Google Search Tools.\footnote{The Google Custom Search API operates on a pay-per-use pricing model. The JSON API, used to retrieve web or image search results, charges \$5 per 1,000 queries. There is a limit of 10,000 queries per day. }}
\vspace{-5pt}
\begin{tcolorbox}[colback=white!95!gray, colframe=black, width=1\textwidth, arc=4mm, boxrule=0.5mm]
\noindent \textbf{The prompt template in Parser:} $P_{tp}$\\  \texttt{\color{black} 
\template{Based on the following instruction, help me decide which tools to use and generate the keywords for tool usage.} \\
\template{Please specify the types of descriptions that the recommended items should include. Do not directly recommend specific items. Don’t use numerical numbering for the generated content; you can use bullet points instead.} \\
\keywords{Instruction $X_I$:} I am on the hunt for a book that offers a refreshing break from the daily grind, much like how a filmmaker can offer a respite from the monotony of car-centric society. The narrative should be feel-good and predictable, with a focus on athletic fashions for everyday people, much like how cars have become a ubiquitous part of modern life. The book should offer an entertaining and engaging escape from reality, much like how a filmmaker can transport the audience to another world through the magic of cinema. } \\
\noindent \textbf{The generated query of Google Search API:} $X_{KW}$\\
\texttt{\color{black} 
\highlightreply{feel-good books that offer an escape from reality and focus on athletic fashion for everyday people.} } \\
\noindent \textbf{The response of Google Search API:}\\
\texttt{\color{black} \keywords{Google Search Results:} $X_{EK}$ \\
May 5, 2024 ... She Wrote the First Great Perimenopause Novel. Miranda July is experimenting again — on the page and in her life. Mar 25, 2023 ... Or even if it doesn't fit this vibe, just any book that makes you happy. Can't wait for all of your recs! :). Jul 21, 2022 ... ... feel good about ourselves. Life is a ... The desire for a good and meaningful life that inspires people to commit daily acts of generosity. Oct 3, 2022 ... The philosopher Galen Strawson believes that some people are simply more “episodic” than others; they're fine living day to day, without regard ... Nov 12, 2020 ... Written by a westerner who came to Buddhism later in life, I found it more accessible when I was new to Buddhist books. But yes, it helped me to ... Jun 9, 2015 ... For all avid readers who have been self-medicating with great books their entire lives, it comes as no surprise that reading books can be good for you. It was a way of life that approximated a racial caste system (Woodward, 1974). Jim Crow laws and etiquette were aided by millions of material objects that ... Apr 5, 2020 ... The Great English Seaside Holiday in its heyday, and the beautiful dignity to be found in everyday living, have rarely been captured more \highlightreply{The rest of the information is omitted.}...} \\
\end{tcolorbox}
\clearpage

\textbf{Without the Tools.} To provide a more reproducible experiment results, we do not utilize the Google Search API all the time, instead of utilizing LLMs to output relative knowledge. Beyond Goole Search API, perplexity search\footnote{https://www.perplexity.ai/} is also a well-performance choice ignoring the price. 
 
\begin{tcolorbox}[colback=white!95!gray, colframe=black, width=1\textwidth, arc=4mm, boxrule=0.5mm]
\noindent \textbf{The Prompt Template in Parser:} $P_{tp}$ \\  \texttt{\color{black} \template{Based on the following instruction, assist me in generating relevant knowledge. \\
Please specify the types of descriptions that the recommended items should include. \\
Do not directly recommend specific items. Don’t use numerical numbering for the generated content; you can use bullet points instead.} \\
 \keywords{Instruction $X_I$:}I am on the hunt for a book that offers a refreshing break from the daily grind, much like how a filmmaker can offer a respite from the monotony of car-centric society. The narrative should be feel-good and predictable, with a focus on athletic fashions for everyday people, much like how cars have become a ubiquitous part of modern life. The book should offer an entertaining and engaging escape from reality, much like how a filmmaker can transport the audience to another world through the magic of cinema.} \\
\noindent \textbf{The Corresponding Response:}\\
\texttt{\color{black} \keywords{LLM Generates Knowledge is:} $X_{IK}$ \\
- Look for a book that features an uplifting and light-hearted storyline, where characters find joy in their daily routines, creating a sense of comfort and familiarity. \\
- Seek narratives that emphasize the importance of community and friendship, perhaps showcasing a diverse cast that showcases various athletic fashions and lifestyles, cheerfully blending sports with daily life. \\
- Consider selections that utilize humor and warmth throughout the narrative, allowing readers to enjoy predictable plot twists that evoke feelings of nostalgia and contentment, similar to classic feel-good films. \\
- Explore themes revolving around self-acceptance and empowerment in the context of athletic fashion, where characters confidently navigate their own styles while participating in community events or activities. \\
- Identify tales that incorporate scenic descriptions of urban or rural settings, inviting readers to visualize a vibrant world beyond their own, akin to the vivid storytelling found in cinema. \\
- Make sure the story maintains a reassuring tone, providing an optimistic perspective on life's challenges, akin to the feel-good resolutions often found in popular films.} \\
\end{tcolorbox}

\subsubsection{Reranker}
\vspace{-10pt}
\begin{tcolorbox}[colback=white!95!gray, colframe=black, width=1\textwidth, arc=4mm, boxrule=0.5mm]
\noindent \textbf{The Prompt Template in Reranker:} $P_{tr}$\\  \texttt{\color{black} \template{Based on the information, give recommendations for the user based on the constrains. Don’t use numerical numbering for the generated content; you can use bullet points instead.}  \\
\keywords{Candidate Ranking List $X_{Item}$:} item id:96578, corresponding title:Surrender, Dorothy: A Novel, description:["Elle Devastatingly on target.The New York Times  ;item id:10837, corresponding title:The Block (Urban Books), description:[''] ;item id:58215, corresponding title:Ritual: A Very Short Introduction (Very Short Intr, description:["Barry Stephenson is Assistant Professor of Relig ;item id:74947, corresponding title:The Collins Case (Heartfelt Cases) (Volume 1), description:['Julie C. Gilbert enjoys writing science fiction, ;item id:173346, corresponding title:Love Handles (A Romantic Comedy) (Oakland Hills), description:['Gretchen Galway is a USA TODAY bestselling autho ;item id:66448, corresponding title:Much Laughter, A Few Tears: Memoirs Of A WomanS Fr, description:[''] ;item id:174617, corresponding title:Drinking at the Movies, description:['', 'Lizzy Caplan Reviews Drinking at the Movies' ;item id:37955, corresponding title:Eternal Now (scm classics), description:["These 16 sermons contain in concentrated form so ;item id:59337, corresponding title:The Guy to Be Seen With, description:["Coming from two generations of journalists, writ ;item id:110713, corresponding title:A Merry Little Christmas: Songs of the Season, description:["Anita Higman is the award-winning author of more , \\
\keywords{Knowledge:{Above Generated Knowledge}},
\keywords{Static Interest $X_{SU}$:}{user historical information, item title:The Executive's Decision: The Keller Family Series,item description:. She is a member of Romance Writers of America and Colorado Romance Writers. Visit her website at www.bernadettemarie.com for news on upcoming releases, signings, appearances, and contests.', '', ''] ;user historical information, item title:Gumbeaux,item description: instructional design content for Fortune 100 companies. Her book, Gumbeaux, received top honors in the 2011 Readers Favorite fiction contest. She lives in San Diego county with her husband Michael.'] ;user historical information, item title:The Hummingbird Wizard (The Annie Szabo Mystery Series) (Volume 1),item description:['', ''] ;user historical information, item title:Artifacts (Faye Longchamp Mysteries, No. 1),item description:['', ''] ;user historical information, item title:3 Sleuths, 2 Dogs, 1 Murder: A Sleuth Sisters Mystery (The Sleuth Sisters) (Volume 2),item description:['Maggie Pill is a lot like Peg Herring, only much cooler and more interesting.'] ;
...(\highlightreply{Pruning.})}, }
\end{tcolorbox}
\vspace{-15pt}
\begin{tcolorbox}[colback=white!95!gray, colframe=black, width=1\textwidth, arc=4mm, boxrule=0.5mm]
\noindent \textbf{The corresponding response:}\\
\texttt{\color{black} 
\keywords{Reranking Results $R^*$:} \{173346, 96578, 110713, 10837, 66448, 74947, 58215, 59337, 174617, 37955\} \\
\noindent \textbf{Answer:} 173346 \\
\noindent \textbf{Initial Ranking List $R$:} [ 96578  10837  58215  74947 173346  66448 174617  37955  59337 110713] 
}
\end{tcolorbox}

\subsubsection{Self-reflection}
Rerank list is out of the order, you should rerank the item from the pure ranking list. The previous list:{}. Therefore, try it again according the following information.
% \vspace{-5pt}
\begin{tcolorbox}[colback=white!95!gray, colframe=black, width=1\textwidth, arc=4mm, boxrule=0.5mm]
\noindent \textbf{Extra Prompt $P_{sr}$:}\\
\texttt{\color{black}
Rerank list is out of the order, you should rerank the item from the pure ranking list. \keywords{The previous list:} \highlightreply{Pure Ranking List}. Therefore, try it again according the following information.  \\
\highlightreply{The rest content is same as the content in the reranker.} \\
}
\end{tcolorbox}

\clearpage

\subsection{Prompt Templates and Examples Response in $\modelplus$}
\subsubsection{Generator}
\begin{tcolorbox}[colback=white!95!gray, colframe=black, width=1\textwidth, arc=4mm, boxrule=0.5mm]
\noindent \textbf{\highlightreply{Step One}} \\
\noindent \textbf{The Prompt Template in Generator $P_{pr1}$:}\\  \texttt{\color{black} \template{Here is the background of one user. } \\
\keywords{Static Interest} \\
Please recommend one item for her. The first one title: Sidewinder (The Halcyon Files Book 1) eBook, descrition:[Nan]. The second one title:Sound Design: The Expressive Power of Music, Voice and Sound Effects in Cinema, description:[Nan].} \\

\noindent \textbf{The Corresponding Response:}\\
\texttt{\color{black} \keywords{Recommended Item $X_G^T$} Based on the provided options and considering a general interest in storytelling and cinematic experiences, I recommend "Sound Design: The Expressive Power of Music, Voice and Sound Effects in Cinema".} \\
\texttt{\color{black} This book delves into the role of sound in film, exploring how music and sound effects enhance narrative and emotional depth. It could provide valuable insights into the art of storytelling through sound, making it a captivating read for anyone interested in filmmaking or audio artistry.}
\\
\noindent \textbf{\highlightreply{Step Two}} \\
\noindent \textbf{The Prompt Template in Generator $P_{pr2}$:}\\  
\texttt{\color{black} Great! Actually, this user choose the item with \highlightreply{title: Sidewinder (The Halcyon Files Book 1) eBook} and \highlightreply{give the review:} lots of twists and turns. Can you generate the profile of this user background? Please make a detailed profile. Don’t use numerical numbering for the generated content; you can use bullet points instead.
. } \\
\noindent \textbf{The Corresponding Response:}\\
\keywords{Generate Profile is $\mathcal{F}^T$:} \texttt{\color{black} - **Name:** [Not Provided] \\
- **Reading Interests:** Enjoys engaging and complex narratives, particularly those with unexpected twists and turns. \\
- **Preferred Genres:** Likely enjoys thriller, mystery, and possibly science fiction or fantasy, given the title of the chosen book, "Sidewinder (The Halcyon Files Book 1)". \\
- **Favorite Themes:** Thrill-seeking adventures, intricate plots, and character development that surprises or challenges expectations. \\
- **Typical Reading Habits:** Prefers books that keep them guessing, possibly reading at night or during quiet moments to fully immerse in the story. \\
- **Current Literary Explorations:** Currently venturing into series that promise suspense and depth, likely seeking to uncover hidden layers within the narrative. \\
- **Community Engagement:** Might engage with online book clubs or forums to discuss their favorite twists in literature and share recommendations. \\}
\end{tcolorbox}
\clearpage
\subsubsection{Extractor}
\begin{tcolorbox}[colback=white!95!gray, colframe=black, width=1\textwidth, arc=4mm, boxrule=0.5mm]
\noindent \textbf{The Prompt Template in Generator:}\\  \texttt{\color{black} \template{Based on the generated knowledge and the instruction, extract some dynamic interest information from the static memory.}  \\
\template{Moreover, based on the profile and the instruction, extract some dynamic profile information. } \\
Don’t use numerical numbering for the generated content; you can use bullet points instead. \\ 
\keywords{Generated Knowledge}:\highlightreply{Knowledge}, \\ \keywords{Instruction}:\highlightreply{Instruction} \\
\keywords{Static Memory}:\highlightreply{Static Interest} \\
\keywords{Profile}:\highlightreply{Generated Profile} .} \\
\noindent \textbf{The Corresponding Response:}\\
\texttt{\color{black} 
\keywords{Dynamic Interest $X_{DU}$:}  \\
- Look for uplifting narrations that are light-hearted and similar to feel-good films, featuring characters finding joy in simple outcomes. \\
- Seek books focused on themes of community and friendship, where relationships flourish in relatable contexts, akin to feel-good cinematic experiences.  \\
- Explore narratives set in vibrant, adventurous locations, enhancing the escapist feel of the story, reminiscent of compelling cinematography. \\
- Consider stories that integrate elements of athletic fashion, reflecting everyday expressions parallel to vehicles' cultural significance in modern life. \\
- Ensure a predictable storyline for easy immersion, providing comfort much like traditional movie endings. \\
- Look for a blend of humor and warmth that echoes the spirit of light-hearted comedies, making them enjoyable reads. \\
- Investigate books featuring authentic characters who embrace their imperfections, promoting positive self-image like many cinematic heroes.  \\
\keywords{Dynamic Profile $\mathcal{F}_d^T$}: \\
- **Reading Interests:** Enjoys feel-good narratives with a focus on community and humor. \\
- **Preferred Genres:** Likely leans towards contemporary fiction, lighthearted romance, and narratives that offer escape and positivity. \\
- **Favorite Themes:** Appreciates uplifting tales that emphasize personal growth, camaraderie, and authenticity in character portrayals. \\
- **Typical Reading Habits:** Prefers reading during relaxation times, seeking stories that provide a comforting escape from daily life, similar to watching feel-good films. \\
- **Current Literary Explorations:** Interested in finding relaxed reads that blend fun and heartwarming elements, potentially exploring themes surrounding fashion and personal identity in a light context. \\
- **Community Engagement:** May participate in discussions around light literature and feel-good storytelling, possibly through book clubs or social media platforms. \\
}
\end{tcolorbox}
\clearpage
\subsection{Examples of Dataset}
\label{appendix::dataset}

\subsubsection{Examples in Constructing Dataset}
\label{appendix::examplesconstruct}
\begin{tcolorbox}[colback=white!95!gray, colframe=black, width=1\textwidth, arc=4mm, boxrule=0.5mm]
\noindent \textbf{The Prompt Template in Constructing Dataset:}\\  \texttt{\color{black} \template{Given the user's review of an item, please mimic the user's instruction which accurately describes their needs.  \\
When crafting each instruction, please make a conscious effort to incorporate a distinct action word or descriptive term that diverges from those showcased in the provided examples.  \\
The reply content should follow the structure: Review text: Persona: Final Instruction: . You should give the initial instruction first based on the reviews and then polish the instruction via mocking the provided persona. But do not reveal the persona directly, just mock their potential writing style. Please provide the instruction based on the review text and decide whether the generated instruction can be used in the examples. \\
Here are some examples.. } \\
Don’t use numerical numbering for the generated content; you can use bullet points instead. \\ 
\keywords{1st Reviews Example}:     Keith Green was a pioneer in the field of Christian rock, and I have loved every album he did. This one is particularly sweet as he was just coming into his own as a premier music writer and performer when it was published. His loss was a terrible blow for millions of his fans. \\
\keywords{1st Personas Example}: A music industry professional with a keen interest in developing new platforms for learning. \\
\keywords{1st Instruction Example}: I'm looking for an exceptional Christian rock album by Keith Green, especially one that showcases his emergence as a premier music writer and performer. His music has a special place in my heart, and something from his prime would be ideal. \\
\keywords{2nd Reviews Example}: I enjoyed the portraits of the heroine going through different transformations: the village girl to the servant to the prostitute to the library clerk...The novel seemed like a picaresque novel from the point of view of an Indian woman: sort of a mash-up of The Little Princess with Vanity Fair. The Pom to Sara to Pamela to Kamala roller coaster starts to become unbelievable towards the end, as the author doesn't spend as much time with the hero's transformation from colonialist to open-hearted husband. \\
\keywords{2nd Personas Example}:A data-driven finance officer responsible for allocating the school district's annual budget. \\
\keywords{2nd Instruction Example}: Seeking a novel that vividly portrays a heroine's transformative journey through various roles, akin to a picaresque tale from an Indian woman's perspective, blending elements of The Little Princess and Vanity Fair. Preferably, the narrative should effectively balance the heroine's evolution with the hero's significant transformation, exploring themes of power dynamics and their impact on relationships. \\
\highlightreply{Other few-shot examples.} \\
\keywords{The User's Review:}} \\
\end{tcolorbox}

\subsubsection{Examples of Filtered Instructions}
\label{appendix:filter}
We use an LLM to filter out instructions that may lead to data leakage. The following examples illustrate some of the filtered instructions.

\begin{tcolorbox}[colback=white!95!gray, colframe=black, width=1\textwidth, arc=4mm, boxrule=0.5mm]
\noindent \textbf{Some Filtered Instructions Examples:}\\  \texttt{\color{black}
\highlightreply{1st example:} As a ticket vendor, I am always on the lookout for a fascinating read that can provide a break from the routine, much like how I seek out the latest comedy films for a good laugh. A book that offers a detailed look into WW2 submarine construction is what I crave. However, I seek a book with clear and detailed photos and drawings, allowing me to fully appreciate the subject matter. The book should be as captivating as a great comedy, providing a mix of entertainment and insight. And just like how I appreciate a good joke, I seek a book that offers a satisfying read, leaving me feeling entertained and informed. The book should leave me feeling like I have learned something new, much like how a successful comedy film can leave a ticket vendor feeling accomplished and motivated to recommend it to others. \\
\highlightreply{2nd example:} In search of a book that offers a comprehensive and insightful look at the genre of mystery novels, much like how a dedicated science blogger can appreciate the intricacies of conducting precise experiments, I seek a narrative that captures the essence of the genre. The book should offer a fresh perspective on the history and evolution of mystery novels, providing a realistic and engaging portrayal of the genre's development. The narrative should be well-written and immersive, offering a depth and complexity that rivals the intricacies of conducting scientific experiments. The book should also offer a nuanced exploration of the challenges and rewards of writing mystery novels, much like how a science blogger can delve into the intricacies of their field of study. \\
\highlightreply{3rd example:} In my search for a book that can offer a fresh and insightful perspective on personality types and relationships, much like how a college professor recovering from a major accident can appreciate the value of alternative medicine, I seek a narrative that can challenge my assumptions and broaden my horizons. The book should offer a well-researched and thoughtful analysis of personality types, much like how a college professor can appreciate the value of evidence-based research. The author should also provide a sense of connection and understanding, much like how a college professor can find value in the human experience and the importance of relationships. A book that meets these criteria would be a valuable addition to any reader's collection, offering a rich and rewarding reading experience that can inspire and inform. \\}

\end{tcolorbox}
\clearpage
\subsubsection{Examples of Retained Instructions}
The following examples show the retained instructions.
\begin{tcolorbox}[colback=white!95!gray, colframe=black, width=1\textwidth, arc=4mm, boxrule=0.5mm]
\noindent \textbf{Some Retained Instructions Examples:}\\  \texttt{\color{black}
\highlightreply{1st example:} In my search for a book that offers a well-researched and informative narrative, much like how a child development researcher can appreciate the nuances of a well-written story that offers accurate and evidence-based information, I seek a resource that offers a comprehensive and engaging look at the subject matter. The book should feature a well-crafted plot that offers a rich history and background, much like how a child development researcher can appreciate the intricacies of a well-written story that offers accurate and evidence-based information. In short, I am seeking a book that offers a comprehensive and informative reading experience, much like how a child development researcher can appreciate the nuances of a well-written story that offers accurate and evidence-based information. \\
\highlightreply{2nd example:} In my search for a book that offers a source of motivation and inspiration, much like how a fellow naval officer with a strong background in logistics and supply chain management collaborates with a young officer on various projects to achieve success, I seek a narrative that can provide a compelling reading experience. The book should be a well-worn companion, offering insights and strategies for building and maintaining a successful career. The writing should be clear and concise, offering a reading experience that is as supportive as a mentor's guidance. And the narrative should offer a balance of action and introspection, much like how a naval officer seeks to balance the practical aspects of their work with a deeper understanding of the complexities and challenges of achieving success. The overall experience should be informative and thought-provoking, much like how a naval officer seeks to gain a deeper understanding of the challenges and opportunities of their career. \\
\highlightreply{3rd example:} In my pursuit of a book that offers a comprehensive guide to business continuity strategies, much like how a strategic planner approaches their work with precision and attention to detail, I seek a narrative that covers all aspects of planning and implementation. The book should be a source of guidance for those who seek to protect their organization from unexpected disruptions, offering a detailed examination of the latest techniques and approaches for ensuring business continuity. A book that meets these criteria would be a valuable addition to my collection, offering a thought-provoking and engaging read that can be enjoyed again and again. However, I request that the list provided to me be accurate and up-to-date, and that any books received in error be returned promptly and without hassle. \\}

\end{tcolorbox}

\section{Future Directions}~\label{sec:future work}

\subsection{More Effective Reranker.} In this version of $\model$ and $\modelplus$, we construct a zero-shot reranker based on LLMs, such as GPT4-o-mini. Recently, several open-source LLMs~\citep{apple2.5b,phi3.8b,gemma2b}, typically containing fewer model parameters (2-3 billion), have demonstrated strong performance. It is feasible to fine-tune smaller LLMs to build a more effective reranker on our $\dataset$ dataset.
Furthermore, existing advanced recommendation models~\citep{metaRec,slmrec} can serve as tools for the agent to retrieve candidate items.
% enhance the reranker's performance. The key challenge lies in integrating collaborative information from these models with the language-based reranker to provide personalized services while reducing interference from other users.

% 1. we have data slm performance well, finetune/ peft build a reranker. 
% 2. Regard rec model as tools combine 
\subsection{Multi-step Feedback.} 

Although we have constructed various datasets rich in abundant instructions, the feedback for re-ranking results is limited to a single ground-truth item, lacking continuous, multi-step feedback on interactions between users and agents. Additionally, the feedback explanations from users are insufficient. If $\modelplus$ were deployed in a real-world environment, more comprehensive feedback could be collected, enabling the development of more interpretable agents for users.

\subsection{Mutual Learning.} This work builds an agent for users that makes decisions for users and collect feedback from users. The platform-side recommendation models can improve their performance by leveraging the feedback and explanations provided by agents on behalf of their users. Furthermore, recommendation agents~\citep{toolrec,agentcf,wang2023recmind,generativerecagent} can autonomously and iteratively improve through mutual learning with $\modelplus$. Moreover, $\modelplus$ can serve as a reward function for RL-based recommendation models~\citep{rl-rec,rl-rec2,rl-rec3,rl-rec4}, enhancing their performance.

\end{document}